\documentclass[10pt,twocolumn,letterpaper]{article}

\usepackage{iccv}
\usepackage{times}
\usepackage{epsfig}
\usepackage{graphicx}
\usepackage{amsmath}
\usepackage{amssymb}
\usepackage{multirow}
\usepackage{floatrow}
\usepackage{anyfontsize}
\usepackage{sidecap}
\newfloatcommand{capbtabbox}{table}[][\FBwidth]
\usepackage{floatrow}

\usepackage[pagebackref=true,breaklinks=true,letterpaper=true,colorlinks,bookmarks=false]{hyperref}
\iccvfinalcopy 

\def\iccvPaperID{1535} 

\begin{document}

\title{Meta R-CNN : Towards General Solver for Instance-level Few-shot Learning}

\author{
		Xiaopeng Yan$^{1}$\thanks{indicate equal contribution (Xiaopeng Yan and Ziliang Chen). $\dagger$ indicates corresponding author: Liang Lin. }, \ \ Ziliang Chen$^{1*}$, \ Anni Xu$^{1}$, \ Xiaoxi Wang$^{1}$, \ Xiaodan Liang$^{1,2}$, \ Liang Lin$^{1,2,\dagger}$\\ $^1$\begin{small}
			Sun Yat-sen University  \ \ $^2$DarkMatter AI Research
			\end{small}\\
	\tt\scriptsize \{yanxp3,wangxx35\}@mail2.sysu.edu.cn, \tt\scriptsize c.ziliang@yahoo.com, \tt\scriptsize 466783266@qq.com, \tt\scriptsize xdliang328@gmail.com, \tt\scriptsize linliang@ieee.org 
}
\maketitle

\begin{abstract}	
Resembling the rapid learning \hspace{-0.1em}capability of human, few-shot learning empowers vision\hspace{-0.1em} systems to understand\hspace{-0.1em} new concepts by \hspace{-0.1em}training \hspace{-0.1em}with few samples. Leading approaches derived from meta-learning on images with a single visual object. Obfuscated by a complex background and multiple objects in one image, they are hard to promote the research of few-shot object detection/segmentation. 
In this work, we present a flexible and general methodology to achieve these tasks. Our work extends Faster /Mask R-CNN by proposing meta-learning over RoI (Region-of-Interest) features instead of a full image feature. This simple spirit disentangles multi-object information merged with the background, without bells and whistles, enabling Faster /Mask R-CNN turn into a meta-learner to achieve the tasks. Specifically, we introduce a Predictor-head Remodeling Network (PRN) that shares its main backbone with Faster /Mask R-CNN. PRN receives images containing few-shot objects with their bounding boxes or masks to infer their class attentive vectors. The vectors take channel-wise soft-attention on RoI features, remodeling those R-CNN predictor heads to detect or segment the objects that are consistent with the classes these vectors represent. In our experiments, Meta R-CNN \hspace{-0.1em}yields the \hspace{-0.1em} state\hspace{-0.1em} of the\hspace{-0.1em} art in \hspace{-0.1em}few-shot \hspace{-0.05em}object detection and improves few-shot object segmentation by \hspace{-0.05em}Mask \hspace{-0.05em}R-CNN.\hspace{-0.1em} Code: \url{https://yanxp.github.io/metarcnn.html}. 
\end{abstract}

\section{Introduction}
\begin{figure}
	\centering \includegraphics[width=8.5cm]{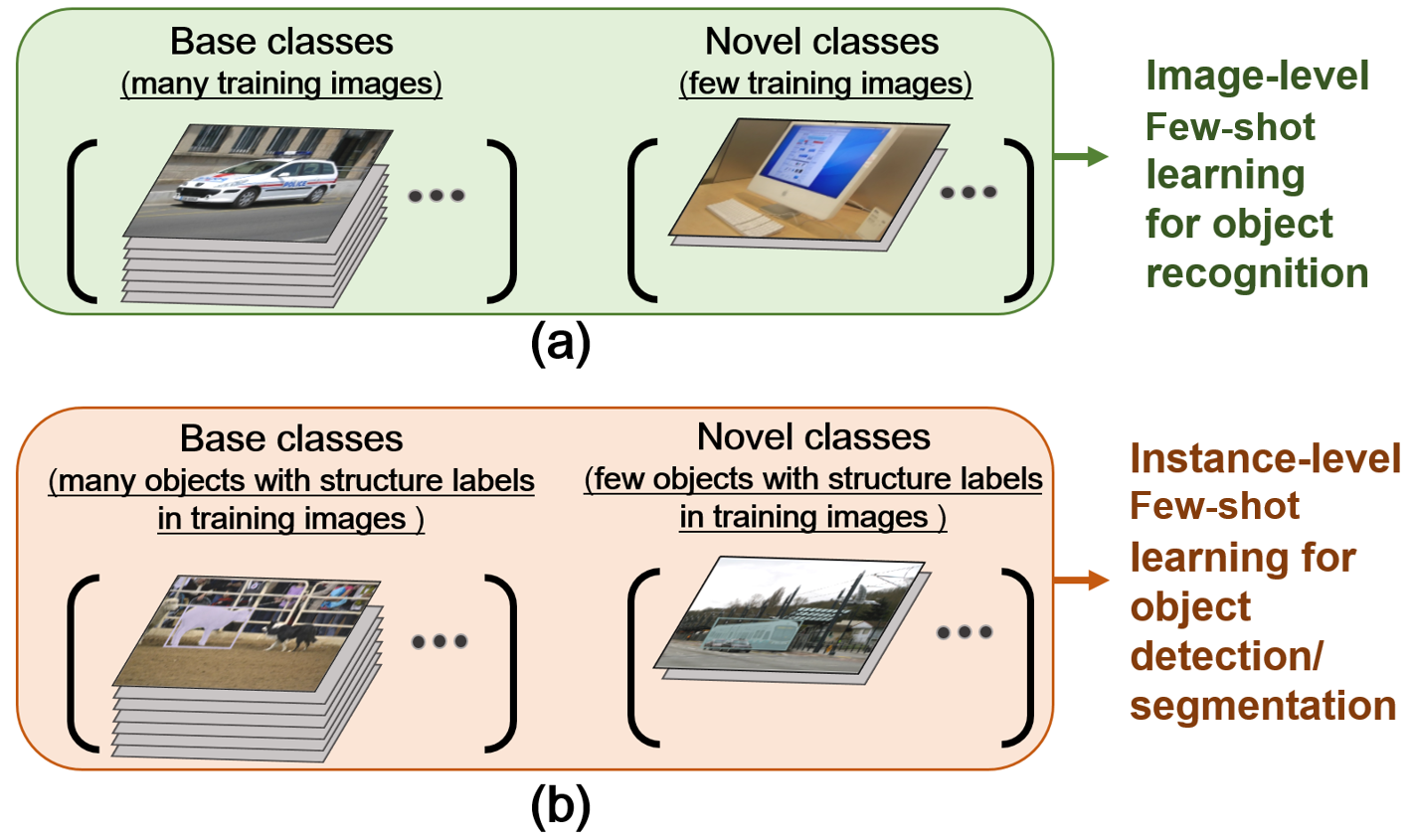} \caption{The illustration of labeled training images in few-shot setups for visual object recognition and class-aware object structure (bounding-boxs or masks) prediction. Compared with recognition, novel-class few objects in few-shot object detection/ segmentation blend with other objects in diverse backgrounds, yet requiring a few-shot learner to predict their classes and structure labels.}\vspace{-0pt}\label{lsl} 
\end{figure}

Deep learning frameworks dominate the vision community to date, due to their human-level achievements in supervised training regimes with a large amount of data. But distinguished with human that excel in rapidly understanding visual characteristics with few demonstrations, deep neural networks significantly \hspace{-0.1em}suffer\hspace{-0.1em} performance drop when training data are scarce in a class. The exposed bottleneck triggers many researches that rethink the generalization of deep learning \cite{Zhang2017Understanding,finn2017model}, among which \emph{few(low)-shot learning} \cite{Lake2015Human} is a popular and very promising direction. Provided with very \emph{few labeled data} (1$\sim$10 shots) in \emph{novel classes}, few-shot learners are trained to recognize the data-starve-class objects by the aid of \emph{base classes} with \emph{sufficient labeled data} (See Fig \ref{lsl}.a). Its industrial potential increasingly drives the emergence of solution, falling under the umbrellas of Bayesian approaches \cite{Fei2008A,Lake2015Human}, similarity learning \cite{koch2015siamese,Shyam2017Attentive} and meta-learning \cite{Vinyals2016Matching,Wang2016Learning,Wang2018Low,Snell2017Prototypical}.

However, \hspace{-0.1em}recognizing \hspace{-0.1em}a single object\hspace{-0.1em} in\hspace{-0.1em} an\hspace{-0.1em} image is solely a tip of the iceberg in real-world visual understanding. In terms of \emph{instance-level learning tasks}, \emph{e.g.}, object detection \cite{Ren2015Faster,redmon2016you}/ segmentation \cite{arnab2016bottom}, prior works in few-shot learning contexts remain rarely explored (See Fig \ref{lsl}.b). Since learning the instance-level tasks requires bounding-box or masks (structure labels) consuming more labors than image-level annotations, it would be practically impactful if the novel classes, object bounding boxes and segmentation masks can be synchronously predicted by a few-shot learner. Unfortunately, these tasks in object-starve conditions become much tougher, as a learner needs to locate or segment the novel-class number-rare objects beside of classifying them. Moreover, due to multiple objects in one image, novel-class objects might blend with the objects in other classes, further obfuscating the information to predict their structure labels. Given this, researchers might expect a complicated solution, as what were done to solve few-shot recognition \cite{Fei2008A,Lake2015Human}.

Beyond their expectation, we present a intuitive and general methodology to achieve few-shot object detection and segmentation : we propose a novel \emph{meta-learning paradigm based on the RoI (Region-of-Interest) features produced by Faster/Mask R-CNN} \cite{Ren2015Faster,He2017Mask}. Faster /Mask R-CNN should be trained with considerable labeled objects and unsuited in few-shot object detection. Existing meta-learning techniques are powerful in few-shot recognition, whereas their successes are mostly based on recognizing a single object. Given an image with multi-object information merged in background, they almost fail as the meta-optimization could not disentangle this complex information. But interestingly, we found that the blended undiscovered objects could be ``pre-processed'' via the RoI features produced by the first-stage inference in Faster /Mask R-CNNs. Each RoI feature refers to a single object or background, so Faster /Mask R-CNN may disentangle the complex information that most meta-learners suffer from. 

Our observation motivates the marriage between Faster /Mask R-CNN and meta-learning. Concretely, we extend Faster /Mask R-CNN by introducing a Predictor-head Remodeling Network (PRN). PRN is fully-convoluted and shares the main backbone's parameters with Faster /Mask R-CNN. Distinct from the R-CNN counterpart, PRN receives few-shot objects drawn from base and novel classes with their bboxes or masks, inferring class-attentive vectors, corresponding to the classes that few-shot input objects belong to. Each vector takes channel-wise attention to all RoI features, inducing the detection or segmentation prediction for the classes. To this end, a Faster /Mask R-CNN predictor head has been remodeled to detect or segment the objects that refer to the PRN's inputs, including the category, position, and structure information of few-shot objects. Our framework exactly boils down to a typical meta-learning paradigm, encouraging the name \emph{Meta R-CNN}. 



Meta R-CNN is \textbf{general} (available in diverse backbones in Faster/Mask R-CNN), \textbf{simple} (a lightweight PRN) yet \textbf{effective} (a huge performance gain in few-shot object detection/ segmentation) and remains \textbf{fast inference} (class-attentive vectors could be pre-processed before testing). We conduct the experiments across 3 benchmarks, 3 backbones for few-shot object detection/ segmentation. Meta R-CNN has achieved the new state of the art in few-shot novel-class object detection/ segmentation, and more importantly, kept competitive performance to detect base-class objects. It verifies Meta R-CNN significantly improve the generalization capability of Faster/ Mask R-CNN.  
  
\begin{SCfigure*}
	\vspace{-6pt}\includegraphics[height=2.45in]{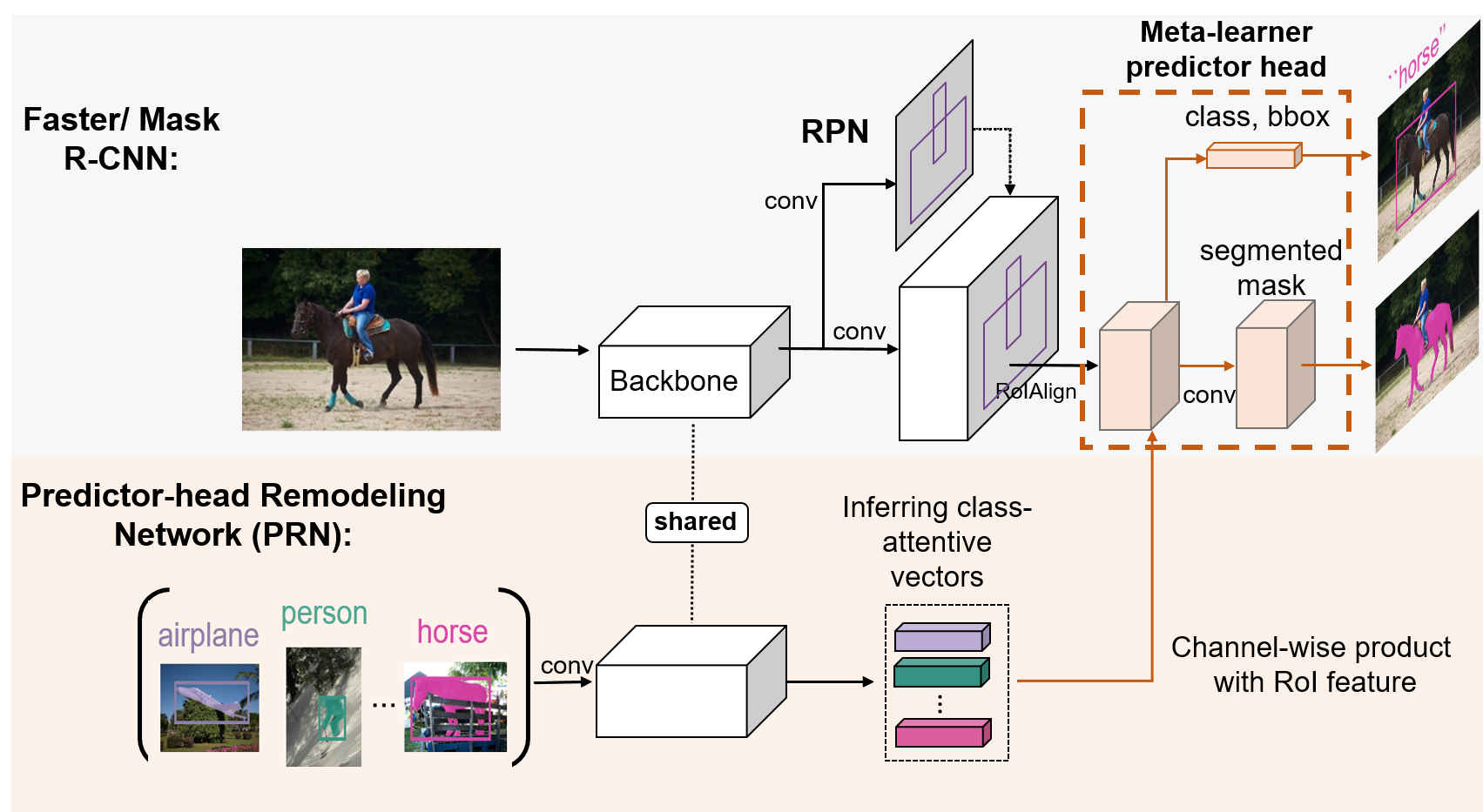}\vspace{-6pt}\label{mrcnn}
	\vspace{14pt}\caption{
		Our Meta R-CNN consists of 1) Faster/ Mask R-CNN; 2) Predictor-head Remodeling Network (PRN). Faster/ Mask R-CNN (module) receives an image to produce RoI features, by taking RoIAlign on the image region proposals extracted by RPN. In parallel, our PRN receives $K$-shot-$m$-class resized images with their structure labels (bounding boxes/segmentaion masks) to infer $m$ class-attentive vectors. Given a class attentive vector representing class $c$, it takes a channel-wise soft-attention on each RoI feature, encouraging the Faster/ Mask R-CNN predictor heads to detect or segment class-$c$ objects based on the RoI features in the image. As the class $c$ is dynamically determined by the inputs of PRN, Meta R-CNN is a meta-learner. 
	}
\end{SCfigure*}

\vspace{-2pt}\section{Related Work}\vspace{-2pt}
\textbf{Few-shot object recognition} aims to recognize novel visual objects given very few corresponding labeled training examples. Recent studies in vision are mainly classed into three streams based on Bayesian approaches, metric learning and meta-learning, respectively. Bayesian approaches \cite{Fei2008A,Lake2015Human} presume a mutual organization rule behind the objects, and design probabilistic model to discover the information among latent variables. Similarity learning~\cite{koch2015siamese,Shyam2017Attentive,sung2018learning} tend to consider the same-category examples's features should be more similar than those between different classes. Distinct from them, meta-learning~\cite{Vinyals2016Matching,Snell2017Prototypical,gidaris2018dynamic,qiao2018few,bertinetto2016learning,ha2016hypernetworks,wang2017learning,finn2017model} designs to learn a \emph{meta-learner} to \emph{parametrize} the optimization algorithm or \emph{predict} the parameters of a classifier, so-called ``learning-to-learn''. Recent theories \cite{amit2017meta,khodak2019provable} show that meta-learner achieves a generalization guarantee, attracting tremendous studies to solve few-shot problems by meta-learning techniques. However, most existing methods focus on single-object recognition.

\textbf{Object detection} based on neural network is mainly resolved by two solver branches: one-stage / two-stage detectors. One-stage detectors attempt to predict bounding boxes and detection confidences of object categories directly, including YOLO ~\cite{redmon2016you}, SSD ~\cite{liu2016ssd} and the variants. R-CNN ~\cite{girshick2014rich} series ~\cite{he2015spatial,girshick2015fast,Ren2015Faster,dai2016r} fall into the second stream. The methods apply covnets to classify and regress the location by the region proposals generated by different algorithms ~\cite{uijlings2013selective,Ren2015Faster}. More recently, few-shot object detection has been extended from recognition~\cite{chen2018lstd,karlinsky2019repmet,kang2018few}. ~\cite{kang2018few} follows full-image meta-learning principle to address this problem. Instead, we discuss the similarity and difference between few-shot object recognition and detection in Sec \ref{3.1}, to reasonably motivate our RoI meta-learning approach. 

\textbf{Object segmentation }is expected to pixel-wise segment the objects of interest in an image. Leading methods are categorized into image-based and proposal-based. Proposal-based methods ~\cite{pinheiro2015learning,pinheiro2016learning,dai2016instance,chen2018masklab} predict object masks based on the generated region proposals while image-based methods ~\cite{zhang2016instance,zhang2015monocular,wu2016bridging,arnab2016bottom} produce a pixel-level segmentation map over the image to identify object instance. The relevant researches in few-shot setup remain absent.

\vspace{-2pt}\section{Tasks and Motivation}\label{3.1}\vspace{-2pt}
Before introducing Meta R-CNN, we consider few-shot object detection /segmentation tasks it aims to achieve. The tasks could be derived from few-shot object recognition in terms of \emph{meta-learning} methods that motivate our method.
 
\vspace{-4pt}\subsection{Preliminary: few-shot visual object recognition by meta-learning}\vspace{-6pt}
In few-shot \hspace{-0.1em}object \hspace{-0.1em}recognition, a \hspace{-0.1em}learner $h(;\boldsymbol{\theta})$ receives training data from \emph{base classes} $C_{\rm base}$ and \emph{novel classes} $C_{\rm novel}$. So the data can be divided into two groups: $D_{\rm base}=\{(\mathbf{x}^{\rm base}_i,y^{\rm base}_i)\}^{n_1}_{i=1}\sim P_{\rm base}$ contains sufficient samples in each base class; 
$D_{\rm novel}=\{(\mathbf{x}^{\rm novel}_i,y^{\rm novel}_i)\}^{n_2}_{i=1}\sim P_{\rm novel}$ contains very few samples in each novel class. $h(;\boldsymbol{\theta})$ aims to classify test samples drawn from 
$P_{\rm novel}$. Notably, training $h(;\boldsymbol{\theta})$ with small dataset  
$D_{\rm novel}$ to identify $C_{\rm novel}$ suffers model overfitting, whereas training $h(;\boldsymbol{\theta})$ with $D_{\rm base}\cup D_{\rm novel}$ still fails, due to the extreme data quantity imbalance between $D_{\rm base}$ and $D_{\rm novel}$ ($n_2<<n_1$). 

Recent wisdoms tend to address this problem by recasting it into a meta-learning paradigm \cite{Vinyals2016Matching,Wang2018Low}, thus, encouraging a fast model adaptation to novel tasks (task generalization), \emph{e.g.}, classifying objects in $C_{\rm novel}$. In each iter, the meta-learning paradigm draws a subset of classes $C_{\rm meta}\sim C_{\rm base}\cup C_{\rm novel}$ and thus, use the images belonging to $C_{\rm meta}$ to construct two batches: a training mini-batch $D_{\rm train}$ and a small-size \emph{meta(reference)-set} $D_{\rm meta}$ (few-shot samples in each class). Given this, a \emph{meta-learner} $h(\mathbf{x}_i, D_{\rm meta};\boldsymbol{\theta})$ simultaneously receives an image $\mathbf{x}_i\sim D_{\rm train}$ and the entire $D_{\rm meta}$ and then, is trained to classify $D_{\rm train}$ into $C_{\rm meta}$\footnote{In a normal setup, meta-learning includes two phases, \emph{meta-train} and \emph{meta-test}. The first phase only use a subset of $C_{\rm base}$ to train a meta-learner.}. By replacing $D_{\rm meta}$ with $D_{\rm novel}$, recent theories \cite{amit2017meta,khodak2019provable} present generalization bounds to the meta-learner, enabling $h(, D_{\rm novel};\boldsymbol{\theta})$ to correctly recognize the objects $\sim P_{\rm novel}$. 

\vspace{-4pt}\subsection{Few-shot object detection / segmentation}\vspace{-6pt}
From visual recognition to detection /segmentation, few-shot learning on objects becomes more complex: An image $\mathbf{x}_i$ might contain $n_i$ objects $\{\mathbf{z}_{i,j}\}^{n_i}_{j=1}$ in diverse classes, positions and shapes. Therefore the few-shot learners need to identify novel-class objects $\mathbf{z}^{\rm novel}_{i,j}$ from other objects and background, and then, predict their classes $y^{\rm novel}_{i,j}$ and structure labels $s^{\rm novel}_{i,j}$ (bounding-boxes or masks). Most existing detection/ segmentation baselines address the problems by modeling $h(\mathbf{x}_i;\boldsymbol{\theta})$, performing poorly in a few-shot scenario. However, meta-predictor $h(\mathbf{x}_i, D_{\rm meta};\boldsymbol{\theta})$ is also unsuitable, since $\mathbf{x}_i$ contains multi-object complex information merged in diverse backgrounds. 

\textbf{Motivation.} The real goal of meta-learning for few-shot object detection/ segmentation is to model $h(\mathbf{z}_{i,j}, D_{\rm meta};\boldsymbol{\theta})$ rather than $h(\mathbf{x}_i, D_{\rm meta};\boldsymbol{\theta})$. Since visual objects $\{\mathbf{z}_{i,j}\}^{n_i}_{j=1}$ are blended with each other and merge with the background in $\mathbf{x}_i$, meta-learning with $\{\mathbf{z}_{i,j}\}^{n_i}_{j=1}$ is intractable. Howbeit in two-stage detection models, \emph{e.g.}, Faster/ Mask R-CNNs,  multi-object and their background information can be disentangled into RoI (Region-of-Interest) features $\{\mathbf{\hat{z}}_{i,j}\}^{\hat{n}_i}_{j=1}$, which are produced by taking RoIAlign on the image region proposals extracted by the region proposal network (RPN). These RoI features are fed into the second-stage predictor head to achieve RoI-based object classification, position location and silhouette segmentation for $\{\mathbf{z}_{i,j}\}^{n_i}_{j=1}$.\hspace{-0.2em} Given this, it is preferable to remodel the R-CNN predictor head into $h(\mathbf{\hat{z}}_{i,j}, D_{\rm meta};\boldsymbol{\theta})$ to classify, locate and segment the object $\mathbf{z}_{i,j}$ behind each region of interest (RoI) feature $\mathbf{\hat{z}}_{i,j}$. 

\vspace{-2pt}\section{Meta R-CNN}\vspace{-2pt}
Aiming at meta-learning over regions of interest (RoIs), Meta R-CNN is conceptually simple: its pipeline consists of 1). Faster/ Mask R-CNN; 2). \emph{Predictor-head Remodeling Network} (\textbf{PRN}). Faster/ Mask R-CNN produces object proposals $\{\mathbf{\hat{z}}_{i,j}\}^{n_i}_{j=1}$ by their region proposal networks (\textbf{RPN}). Then each $\mathbf{\hat{z}}_{i,j}$ combines with \emph{class-attentive vectors} inferred by our \textbf{PRN}, which plays the role of $h(, D_{\rm meta};\boldsymbol{\theta})$ to detect or segment the novel-class objects. The Meta R-CNN framework is illustrated in Fig \ref{mrcnn} and we elaborate it by starting from Faster/ Mask R-CNN. 

\vspace{-4pt}\subsection{Review the R-CNN family}\vspace{-4pt}
Faster R-CNN system is known as a two-stage pipeline. \hspace{-0.4em}The first stage is a region proposal network (\textbf{RPN}), receiving an image $\mathbf{x}_i$ to produce the candidate object bounding-boxes (so-called object region proposals) in this image. The second stage, \emph{i.e.}, Fast R-CNN \cite{girshick2015fast}, shares the RPN backbone to extract RoI (Region-of-Interest) features $\{\mathbf{\hat{z}}_{i,j}\}^{\hat{n}_i}_{j=1}$ from $\hat{n}_i$ object region proposals after RoIAlign \footnote{RoIAlign operation is first introduced by Mask R-CNN yet can be used by Faster R-CNN. Faster R-CNNs in our work are based on RoIAlign.} , enabling its predictor head $h(\mathbf{\hat{z}}_{i,j},\boldsymbol{\theta})$ to classify and locate the object $\mathbf{z}_{i,j}$ behind the RoI feature $\mathbf{\hat{z}}_{i,j}$ of $\mathbf{x}_i$. Mask R-CNN activates the segmentation ability in Faster R-CNN by adding a parallel mask branch in the predictor head $h(\cdot,\boldsymbol{\theta})$. Due to our identical technique applied in Faster/ Mask R-CNN, we unify their predictor heads by $h(\cdot,\boldsymbol{\theta})$.

As previously discussed, predictor head $h(\cdot,\boldsymbol{\theta})$ in Faster/ Mask R-CNN is inappropriate to make few-shot object detection/ segmentation. To this we propose PRN that remodels $h(\cdot,\boldsymbol{\theta})$ into a meta-predictor head $h(\cdot , D_{\rm meta};\boldsymbol{\theta})$.
\vspace{-4pt}\subsection{Predictor-head Remodeling Network (PRN)}\vspace{-6pt}
A straightforward approach to design $h(\cdot , D_{\rm meta};\boldsymbol{\theta})$ is to learn $\boldsymbol{\theta}$ to predict the optimal parameter $\emph{w.r.t.}$ an arbitrary meta-set $D_{\rm meta}$ like \cite{Wang2016Learning}. Such explicit ``learning-to-learn'' manner is sensitive to the architectures and $h(\cdot,\boldsymbol{\theta})$ in Faster/ Mask R-CNN is abandoned. Instead, our work is inspired by the concise spirit of SNAIL \cite{Mishra2017A}, thus, incorporating class-specific soft-attention vectors to achieve channel-wise feature selection on each RoI feature in $\{\mathbf{z}_{i,j}\}^{\hat{n}_i}_{j=1}$ \cite{Chen2016SCA}. This soft-attention mechanism is implemented by the class-attentive vectors $\mathbf{v}^{\rm meta}$ inferred from the objects in a meta-set $D_{\rm meta}$ via PRN. In particular, suppose that PRN denotes as $\mathbf{v}^{\rm meta}=f(D_{\rm meta};\boldsymbol{\phi})$, given each RoI feature $\hat{\mathbf{z}}_{i,j}$ that belongs to image $\mathbf{x}_i$, it holds
\begin{equation}\begin{aligned}
h(\hat{\mathbf{z}}_{i,j} , D_{\rm meta};\boldsymbol{\theta}') &= h(\hat{\mathbf{z}}_{i,j}\otimes\mathbf{v}^{\rm meta} ,\boldsymbol{\theta})\\&=h(\hat{\mathbf{z}}_{i,j}\otimes f(D_{\rm meta};\boldsymbol{\phi}) ,\boldsymbol{\theta})
\end{aligned}\label{metaeq}
\end{equation}where $\boldsymbol{\theta}$, $\boldsymbol{\phi}$ denote the parameters of Faster/Mask R-CNN and our PRN (most of them are shared, $\boldsymbol{\theta}'=\{\boldsymbol{\theta},\boldsymbol{\phi}\}$); $\otimes$ indicates the channel-wise multiplication operator. Eq \ref{metaeq} implies that PRN remodels $h(\cdot,\boldsymbol{\theta})$ into $h(\cdot , D_{\rm meta};\boldsymbol{\theta})$ in principles. It is intuitive, flexibly-applied and allows end-to-end joint training with its Faster/ Mask R-CNN counterpart. 

Suppose $\mathbf{x}_i$ as the image Meta R-CNN aiming to detect. After RoIAlign in its R-CNN module, it turns to a set of RoI features $\{\mathbf{\hat{z}}\}^{\hat{n}_i}_{i,j}$. Here we explain how PRN acts on them.   
 
\textbf{Infer class-attentive vectors.} As can be observed, PRN $f(D_{\rm meta};\boldsymbol{\phi})$ receives all objects in meta-set $D_{\rm meta}$ as input. In the context of object detection/ segmentation, $D_{\rm meta}$ denotes a series of objects distributed across images, whose classes belong to $C_{\rm meta}$ and there exist $K$ objects per class ($K$-shot setup). \hspace{-0.2em}Each object in $D_{\rm meta}$ presents a 4-channel input, \emph{i.e.}, an RGB image $\mathbf{x}$ with the same-spatial-size foreground structure label $s$ that are combined to represent this object ($s$ is a binary mask derived from the object bounding-box or segmentation mask). Hence given $m$ as the size of $C_{\rm meta}$, PRN receives $mK$ 4-channel object inputs in each inference process. To ease the computation burden, we standardize the spatial size of object inputs into 224$\times$224. During inference, after passing the first convolution layer of our PRN, each object feature would be fed into the second layer of its R-CNN counterpart, undergoing the shared backbone before RoIAlign. Instead of accepting RoIAlign, the feature passes a channel-wise soft-attention layer to produce its \emph{object attentive vector} $\mathbf{v}$. To this end, PRN encodes $mK$ objects in $D_{\rm meta}$ into $mK$ object attentive vectors and then, applies average pooling to obtain the class-attentive vectors $\mathbf{v}_{c}^{\rm meta}$, \emph{i.e.},
$
\mathbf{v}_{c}^{\rm meta}= \frac{1}{K}\sum_{j=1}^{K}\mathbf{v}_k^{(c)}, 
$ ($\forall c\in C_{\rm meta}$, $\mathbf{v}_k^{(c)}$ represents an object attentive vector inferred from a class-$c$ object and there are $K$-shot objects per class).  

\textbf{Remodel R-CNN predictor heads.} After obtaining the class-attentive vectors $\mathbf{v}_{c}^{\rm meta}$ ($\forall c\in C_{\rm meta}$), PRN applies them to take channel-wise soft-attention on each RoI feature $\mathbf{z}_{i,j}$. Suppose that $\mathbf{\hat{Z}}_i=[\mathbf{\hat{z}}_{i,1};\cdots;\mathbf{\hat{z}}_{i,128}]\in \mathbb{R}^{2048\times128}$ denotes the RoI feature matrix generated from $\mathbf{x}_i$ ($128$ denotes the number of RoI). PRN replaces $\mathbf{\hat{Z}}_i$ by $\mathbf{\hat{Z}}_i\otimes \mathbf{v}_{c}^{\rm meta}=[\mathbf{\hat{z}}_{i,1}\otimes \mathbf{v}_{c}^{\rm meta};\cdots;\mathbf{\hat{z}}_{i,128}\otimes \mathbf{v}_{c}^{\rm meta}]$ to feed the primitive predictor heads in Faster /Mask R-CNNs. The refinement leads to detecting or segmenting all class-$c$ objects in the image $\mathbf{x}_i$. In this spirit, each RoI feature $\mathbf{\hat{z}}_{i,j}$ produces $m$ binary detection outcomes that refers to the classes in $C_{\rm meta}$. To this Meta R-CNN categorizes $\mathbf{\hat{z}}_{i,j}$ into the class $c^\ast$ with the highest confidence score and use the branch $\mathbf{\hat{z}}_{i,j}\otimes \mathbf{v}_{c^\ast}^{\rm meta}$ to locate or segment the object. But if the highest confidence score is lower than the objectness threshold, this RoI would be treated as background and discarded.

\vspace{-2pt}\section{Implementation}\vspace{-2pt}
Meta R-CNN is trained under a meta-learning paradigm. Our implementation based on Faster/ Mask R-CNN, whose hyper-parameters follow their original report.  

\textbf{Mini-batch construction.} Simulating the meta-learning paradigm we have discussed, a training mini-batch in Meta R-CNN is comprised of $m$ classes $C_{meta}\sim C_{\rm base}\cup C_{\rm novel}$, a $K$-shot $m$-class meta-set $D_{\rm meta}$ and $m$-class training set $D_{\rm train}$ (classes in $D_{\rm meta}$, $D_{\rm train}$ consistent with $C_{meta}$). In our implementation, $D_{\rm train}$ represent the objects in the input $\mathbf{x}$ of Faster/ Mask R-CNNs. To keep the class consistency, we choose $C_{meta}$ as the object classes image $\mathbf{x}$ refers to, and only uses the attentive vectors inferred from the objects belonging to the classes in $C_{\rm meta}$. Therefore, if the R-CNN module receives an image input $\mathbf{x}$ that contains objects in $m$ classes, a mini-batch consists of $\mathbf{x}$ ($D_{\rm train}$) and $mK$ resized images with their structure label masks.  

\textbf{Channel-wise soft-attention layer.} This layer receives the features induced from the main backbone of the R-CNN counterpart. It performs a spatial pooling to align the object features maintaining the identical size of RoI features. Then these features undergo an element-wise sigmoid to produce attentive vectors (the size is 2048$\times$1 in our experiment).  

\textbf{Meta-loss.} Given an RoI feature $\mathbf{\hat{z}}_{i,j}$, to avoid the prediction ambiguity after soft-attention, attentive vectors from different-class objects should lead to diverse feature selection effects\hspace{-0.1em} on $\mathbf{\hat{z}}_{i,j}$\hspace{-0.4em}. To achieve\hspace{-0.1em} this we propose a simple \emph{meta-loss} $L(\boldsymbol{\phi})_{\rm meta}$ to diversify the inferred object attentive vectors in meta-learning. It is implemented by a cross-entropy loss encouraging the object attentive vectors to fall in the class each object belongs to. This auxiliary loss powerfully boosts Meta R-CNN performance (see Table \ref{tab:ablations} Ablation {\color{red}2}).  

\textbf{RoI meta-learning.} Following the typical optimization routines in \cite{Vinyals2016Matching,Snell2017Prototypical,Wang2018Low}, meta-learning Meta R-CNN is divided into two learning phases. In the first phase (so-called \emph{meta-train}), we solely consider base-class objects to construct $D_{\rm meta}$ and $D_{\rm train}$ per iter. In the case that an image simultaneously includes base-class and novel-class objects, we ignore the novel-class objects in meta-train. In the second phase (so-called \emph{meta-test}), objects in base and novel classes are both considered. The objective is formulated as
\begin{equation}
\underset{\boldsymbol{\theta}, \ \boldsymbol{\phi}}{\min} \ \underbrace{L(\boldsymbol{\theta}, \boldsymbol{\phi})_{\rm cls}+L(\boldsymbol{\theta}, \boldsymbol{\phi})_{\rm reg}+\lambda L(\boldsymbol{\theta}, \boldsymbol{\phi})_{\rm mask}}_{\rm Losses \ derived \ from \ Faster/Mask \ R-CNN}+L(\boldsymbol{\phi})_{\rm meta}\label{obj}
\end{equation}where $\lambda =\{0,1\}$ indicates the activator of mask branch. The illustration of meta-learning for Meta R-CNN is below:
\begin{figure}[h]
	\centering
	\vspace{-12pt}\includegraphics[width = 0.9\textwidth]{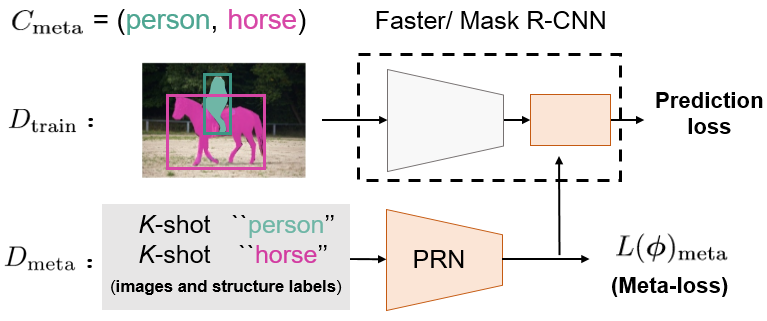}\vspace{2pt}
	\caption{The illustrative instance of meta-optimization process in Meta R-CNN. Suppose the image Faster/ Mask R-CNN receiving contains objects in ``person'', ``horse''. Then $C_{\rm meta}=($ ``person'', ``horse'' $)$ and $D_{\rm meta}$ includes $K$-shot ``person'' and ''horse'' images with their structure labels, respectively. As the training image iteratively changes, $C_{\rm meta}$ and $D_{\rm meta}$ would adaptively change.}
	\label{meta-learning}
	\vspace{-6pt}
\end{figure}

\vspace{-6pt}\textbf{Inference.} Meta R-CNN entails two inference processes based on Faster/Mask R-CNN module and PRN. In training, the object attentive vectors inferred from $D_{\rm meta}$ would replace the class-attentive vectors to take soft-attention effects on $\mathbf{\hat{Z}}_i$ and produce the object detection/ segmentation losses in Eq \ref{obj}.  
In testing, we choose $C_{\rm meta}=C_{\rm base}\cup C_{\rm novel}$. It is because that unknown objects in a test image may cover all possible categories. PRN receives $K$-shot visual objects in all classes to produce class-attentive vectors to achieve few-shot object detection/ segmentation. Note that, no matter of object or class attentive vectors, they can be pre-processed before testing, and parallelly take soft-attention on RoI feature matrices. It promises the fast inference of Faster/ Mask R-CNN will not be decelerated: In our experiment (using a single GTX TITAN XP), if shot is 3, the inference speed of Faster R-CNN is 83.0 ms/im; Meta R-CNN is 84.2ms/im;
if shot is 10, the speed of Meta R-CNN is 85.4ms/im.

\begin{table*}[!htb]
	\begin{floatrow}
		\setlength{\tabcolsep}{2pt}
		{\fontsize{8}{9.8}\selectfont
			\capbtabbox{
				\renewcommand\arraystretch{1.3}
				\begin{tabular}{c |c c c c c | c c c c c | c c c c c }
					\hline
					\multicolumn{1}{c|}{}&\multicolumn{5}{c|}{Novel-class Setup 1}&\multicolumn{5}{c|}{Novel-class Setup 2}&\multicolumn{5}{c}{Novel-class Setup 3}\cr\cline{1-16}
					\hline
					Method/Shot & 1 & 2 & 3 & 5 & 10 & 1 & 2 & 3 & 5 & 10 & 1 & 2 & 3 & 5 & 10\cr\hline
					YOLO-Few-shot~\cite{kang2018few} &{\color{blue}\textbf{14.8}} &{\color{blue}\textbf{15.5}} &26.7 &33.9 &{\color{blue}\textbf{47.2}} &{\color{red}\textbf{15.7}} &{\color{blue}\textbf{15.3}} &22.7 &30.1 &{\color{blue}\textbf{39.2}} &{\color{red}\textbf{19.2}} &{\color{red}\textbf{21.7}} &{\color{blue}\textbf{25.7}} &{\color{blue}\textbf{40.6}} &41.3 \\
					FRCN+joint &2.7 &3.1 &4.3 &11.8 &29.0 &1.9 &2.6 &8.1 &9.9 &12.6 &5.2 &7.5 &6.4 &6.4 &6.4 \\
					FRCN+ft &11.9 &16.4 &29.0 &36.9 &36.9 &5.9 &8.5 &23.4 &29.1 &28.8 &5.0 &9.6 &18.1 &30.8 &43.4 \\
					FRCN+ft-full &13.8 &{\color{blue}\textbf{19.6}} &{\color{blue}\textbf{32.8}} &{\color{blue}\textbf{41.5}} &45.6 &7.9 &{\color{blue}\textbf{15.3}} &{\color{blue}\textbf{26.2}} &{\color{blue}\textbf{31.6}} &39.1 &9.8 &11.3 &19.1 &35.0 &{\color{blue}\textbf{45.1}} \\
					Meta R-CNN (ours) &{\color{red}\textbf{19.9}} &{\color{red}\textbf{25.5}} &{\color{red}\textbf{35.0}} &{\color{red}\textbf{45.7}} &{\color{red}\textbf{51.5}} &{\color{blue}\textbf{10.4}} &{\color{red}\textbf{19.4}} &{\color{red}\textbf{29.6}} &{\color{red}\textbf{34.8}} &{\color{red}\textbf{45.4}} &{\color{blue}\textbf{14.3}} &{\color{blue}\textbf{18.2}} &{\color{red}\textbf{27.5}} &{\color{red}\textbf{41.2}} &{\color{red}\textbf{48.1}} \\
					\hline
				\end{tabular}
				
			}{
			\caption{Few-shot detection mAP on VOC2007 test set in novel classes. We evaluate the baselines under three different splits of novel classes. {\color{red}RED} and {\color{blue}BLUE} indicate state-of-the-art (SOTA) and the second best. (Best viewd in color)}\label{tab:novel}
		}
		\capbtabbox{
			\renewcommand\arraystretch{1}
			\begin{tabular}{c|c|c|c}
				\hline
				Shot &Baselines &Base &Novel \cr\hline
				\multirow{4}*{3}
				&ResNet-34+ft-full &\textbf{57.9} &19.6\\
				&ResNet-34+Ours &57.6 &\textbf{25.3}\\
				\cline{2-4}
				&ResNet-101+ft-full &63.6 &32.8 \\ 
				&ResNet-101+Ours &\textbf{64.8} &\textbf{35.0}\\
				\hline
				\multirow{4}*{10}
				&ResNet-34+ft-full &61.1  &40.2\\
				&ResNet-34+Ours &\textbf{61.3}  &\textbf{44.5}\\
				\cline{2-4}
				&ResNet-101+ft-full &61.3 &45.6\\ 
				&ResNet-101+Ours &\textbf{67.9} &\textbf{51.5}\\
				\hline
			\end{tabular}
		}{
		\caption{The ablation study of backbones (mAP on VOC2007 testset in novel classes and base classes of the first base/novel split based on FRCN).
		}\label{tab:resnet}
	}
}
\end{floatrow}
\end{table*}

\begin{table*}[htp]
	\center
	\setlength{\tabcolsep}{1.2pt}
	\fontsize{7.6}{10.5}{\selectfont
		\begin{tabular}{c |c| c c c c c c | c c c c c c c c c c c c c c c c|c}
			\hline
			\multicolumn{2}{c|}{}&\multicolumn{6}{c|}{Novel classes}&\multicolumn{16}{c|}{Base classes}&\multicolumn{1}{c}{\multirow{2}*{mAP}}\cr\cline{1-24}
			Shot & Baselines & bird & bus & cow & mbike & sofa & mean & aero   & bike & boat & bottle & car & cat & chair & table & dog & horse & person & plant & sheep & train & tv 
			& mean \cr\cline{1-23} \hline
			\multirow{5}*{3}
			&YOLO-Few-shot~\cite{kang2018few} &26.1 &19.1 &40.7 &20.4 &\color{red}\textbf{27.1} &26.7 &\color{blue}\textbf{73.6} &\color{blue}\textbf{73.1} &56.7 &41.6 &\color{blue}\textbf{76.1} &78.7 &42.6 &\color{blue}\textbf{66.8} &72.0 &77.7 &68.5 &\color{blue}\textbf{42.0} &57.1 &\color{blue}\textbf{74.7} &70.7 &\color{blue}\textbf{64.8} &55.2 \\
			&FRCN+joint &13.7 &0.4 &6.4 &0.8 &0.2 &4.3 &\color{red}\textbf{75.9} &\color{red}\textbf{80.0} &\color{red}\textbf{65.9} &\color{red}\textbf{61.3} &\color{red}\textbf{85.5} &\color{red}\textbf{86.1} &\color{red}\textbf{54.1} &\color{red}\textbf{68.4} &\color{red}\textbf{83.3} &\color{red}\textbf{79.1} &\color{red}\textbf{78.8} &\color{red}\textbf{43.7} &\color{red}\textbf{72.8} &\color{red}\textbf{80.8} &\color{red}\textbf{74.7} &\color{red}\textbf{72.7} &55.6 \\
			&FRCN+ft &\color{red}\textbf{31.1} &24.9 &\color{blue}\textbf{51.7} &23.5 &13.6 &29.0 &65.4 &56.4 &46.5 &41.5 &73.3 &84.0 &40.2 &55.9 &72.1 &75.6 &74.8 &32.7 &60.4 &71.2 &\color{blue}\textbf{71.2} &61.4 &53.3\\
			&FRCN+ft-full &29.1 &\color{blue}\textbf{34.1} &\color{red}\textbf{55.9} &\color{blue}\textbf{28.6} &\color{blue}\textbf{16.1} &\color{blue}\textbf{32.8} &67.4 &62.0 &54.3 &48.5 &74.0 &\color{blue}\textbf{85.8} &42.2 &58.1 &72.0 &\color{blue}\textbf{77.8} &75.8 &32.3 &\color{blue}\textbf{61.0} &73.7 &68.6 &63.6 &\color{blue}\textbf{55.9}\\
			&Meta R-CNN (ours) &\color{blue}\textbf{30.1} &\color{red}\textbf{44.6} &50.8 &\color{red}\textbf{38.8} &10.7 &\color{red}\textbf{35.0} &67.6 &70.5 &\color{blue}\textbf{59.8} &\color{blue}\textbf{50.0} &75.7 &81.4 &\color{blue}\textbf{44.9} &57.7 &\color{blue}\textbf{76.3} &74.9 &\color{blue}\textbf{76.9} &34.7 &58.7 &\color{blue}\textbf{74.7} &67.8 &\color{blue}\textbf{64.8} &\color{red}\textbf{57.3} \\
			\hline
			\multirow{5}*{10}
			&YOLO-Few-shot~\cite{kang2018few} &30.0 &\color{red}\textbf{62.7} &43.2 &\color{red}\textbf{60.6} &39.6 &\color{blue}\textbf{47.2} &65.3 &73.5 &54.7 &39.5 &75.7 &81.1 &35.3 &62.5 &72.8 &78.8 &68.6 &\color{blue}\textbf{41.5} &59.2 &\color{blue}\textbf{76.2} &69.2 &63.6 &\color{blue}\textbf{59.5}\\
			&FRCN+joint &14.6 &20.3 &19.2 &24.3 &2.2 &16.1 
			&\color{red}\textbf{78.1} &\color{red}\textbf{80.0} &\color{red}\textbf{65.9} &\color{red}\textbf{64.1} &\color{red}\textbf{86.0} &\color{red}\textbf{87.1} &\color{red}\textbf{56.9} &\color{red}\textbf{69.7} &\color{red}\textbf{84.1} &\color{blue}\textbf{80.0} &\color{red}\textbf{78.4} &\color{red}\textbf{44.8} &\color{red}\textbf{74.6} &\color{red}\textbf{82.7} &\color{red}\textbf{74.1} &\color{red}\textbf{73.8} &59.4\\
			&FRCN+ft &31.3 &36.5 &\color{red}\textbf{54.1} &26.5 &36.2 &36.9 &\color{blue}\textbf{68.4} &\color{blue}\textbf{75.2} &59.2 &\color{blue}\textbf{54.8} &74.1 &80.8 &42.8 &56.0 &68.9 &77.8 &75.5 &34.7 &\color{blue}\textbf{66.1} &71.2 &66.2 &64.8 &57.8 \\
			&FRCN+ft-full &\color{blue}\textbf{40.1} &47.8 &45.5 &47.5 &\color{red}\textbf{47.0} &45.6 &65.7 &69.2 &52.6 &46.5 &74.6 &73.6 &40.7 &55.0 &69.3 &73.5 &73.2 &33.8 &56.5 &69.8 &65.1 &61.3 &57.4\\
			&Meta R-CNN (ours) &\color{red}\textbf{52.5} &\color{blue}\textbf{55.9} &\color{blue}\textbf{52.7} &\color{blue}\textbf{54.6} &\color{blue}\textbf{41.6} &\color{red}\textbf{51.5} &68.1 &73.9 &\color{blue}\textbf{59.8} &54.2 &\color{blue}\textbf{80.1} &\color{blue}\textbf{82.9} &\color{blue}\textbf{48.8} &\color{blue}\textbf{62.8} &\color{blue}\textbf{80.1} &\color{red}\textbf{81.4} &\color{blue}\textbf{77.2} &37.2 &65.7 &75.8 &\color{blue}\textbf{70.6} &\color{blue}\textbf{67.9} &\color{red}\textbf{63.8}\\
			\hline
		\end{tabular}}
		\caption{AP and mAP on VOC2007 test set for novel classes and base classes of the first base/novel split. We evaluate the performance for 3/10-shot novel-class examples with FRCN under ResNet-101. {\color{red}RED}/{\color{blue}BLUE} indicate the SOTA/the second best. (Best viewd in color)
		}\label{tab:sub_split1}
		\vspace{-12pt}
	\end{table*}
	
\vspace{-2pt}\section{Experiments}\vspace{-2pt}
In this section, we propose thorough experiments to evaluate Meta R-CNN on few-shot object detection, the related ablation, and few-shot object segmentation. 

\vspace{-0pt}\subsection{Few-shot object detection}\label{6.1}\vspace{-2pt}
In few-shot object detection, we employ a Faster R-CNN ~\cite{Ren2015Faster} with ResNet-101~\cite{He2017Mask} backbone as the R-CNN module in our Meta R-CNN framework. 

\textbf{ Benchmarks and setups. }Our few-shot object detection experiment follows the setup \cite{kang2018few}. Concretely, we evaluate all baselines on the generic object detection tracks of PASCAL VOC 2007~\cite{everingham2010pascal}, 2012~\cite{everingham2010pascal}, and MS-COCO~\cite{lin2014microsoft} benchmarks. We adopt the PASCAL Challenge protocol that a correct prediction should have more than 0.5 IoU with the ground truth and set the evaluation metric to the mean Average Precision (mAP). Among these benchmarks, VOC 2007 and 2012 consists of images covering 20 object categories for training, validation and testing sets. To create a few-shot learning setup, we consider three different novel/base-class split settings, \emph{i.e.}, (``bird'',
``bus'',
``cow'', ``mbike'', ``sofa''/ rest);  (``aero'', ``bottle'',``cow'',``horse'',``sofa'' / rest) and (``boat'',
``cat'',
``mbike'',``sheep'', ``sofa''/ rest). During the first phase of meta-learning, only base-class objects are considered. In the second phase, there are $K$-shot annotated bounding boxes for objects in each novel class and $3K$ annotated bounding boxes for objects in each base class for training, where $K$ is set 1, 2, 3, 5 and 10. We also evaluate our method on COCO benchmark with 80 object categories including the 20 categories in PASCAL VOC. In this experiment, we set the 20 classes included in PASCAL VOC as the novel classes, then the rest 60 classes in COCO as base classes. The union of 80k train images and a 35k subset of validation images (trainval35k) are used for training, and our evaluation is based on the remaining 5k val images (minival). Finally, we consider the cross-benchmark transfer setup of few-shot object detection from \emph{COCO} to \emph{PASCAL} ~\cite{kang2018few}, which leverages 60 base classes of COCO to learn knowledge representations and the evaluation is based on 20 novel classes of PASCAL VOC.

\textbf{Baselines.} In methodology, Meta R-CNN can be treated as the meta-learning extension of Faster R-CNN (FRCN) ~\cite{Ren2015Faster} in the background of few-shot object detection. To this a question about detector generalization is probably raised:

\begin{center}
	\emph{\textbf{Does Meta R-CNN help to improve the generalization capability of Faster R-CNN?}}
\end{center}

To answer this question, we compare our Meta R-CNN with its base FRCN. This detector is derived into three baselines according to the different training strategies they use. Specifically, \textbf{FRCN+joint} is to jointly train the FRCN detector with base-class and novel-class objects. The identical number of iteration is used for training this baseline and our Meta R-CNN. \textbf{FRCN+ft} takes a similar two-phase training strategy in Meta R-CNN: it only uses base-class objects (with bounding boxes) to train FRCN in the first phase, then use the combination of base-class and novel-class objects to fine-tune the network. For a fair comparison, the objects in images used to train FRCN+ft is identical to Meta R-CNN, and FRCN+ft also takes the same number of iteration (in both training phases) of Meta R-CNN. Finally, \textbf{FRCN+ft-full} employ the same training strategy of FRCN+ft in the first phase, yet train the detector to fully converge in the second phase. Beyond these baselines, Meta R-CNN is also compared with the state-of-the-art few-shot object detector~\cite{kang2018few} modified from YOLOv2~\cite{redmon2017yolo9000} (\textbf{YOLO-Few-shot}). Note that, YOLO-Few-shot also employs meta-learning, whereas distinct from Meta R-CNN based on RoI features, it is based on a full image. Their comparison reveals whether the motivation of Meta R-CNN is reasonable.

\textbf{PASCAL VOC.} The experimental evaluation are shown in Table {\color{red}\ref{tab:novel}}. The $K$-shot object detection is performed based on $K=(1,2,3,5,10)$ across three novel/base class splits. As can be observed, Meta R-CNN consistently outperforms the three FRCN baselines by a large margin across splits. It uncovers the generalization weakness of FRCN: without adequate number of bounding-box annotations, FRCN performs poorly to detect novel-class objects, and this weakness could not be overcome by changing the training strategies. In a comparison, by simply deploying a lightweight PRN, FRCN turns into Meta R-CNN and significantly improve the performance on novel-class object detection. It implies that our approach endows FRCN with the generalization ability in few-shot learning. 

Besides, Meta R-CNN outperforms YOLO-Few-shot in the majority of the cases (except for 1/2-shot in the third split). Since the YOLO-Few-shot results are borrowed from their report, the 1/2-shot objects are probably different from what we use. Extremely-few-shot setups are sensitive to the change of the few-shot object selection and thus, hard to reveal the superiority of few-shot learning algorithms. In the more robust 5/10-shot setups, Meta R-CNN significantly exceeds YOLO-Few-shot ($+11.8\%$ in the 5-shot of the first split; $+6.8$ in the 10-shot of the third split.) 

\begin{figure*}[htb]
	\centering
	\includegraphics[width = 1\textwidth]{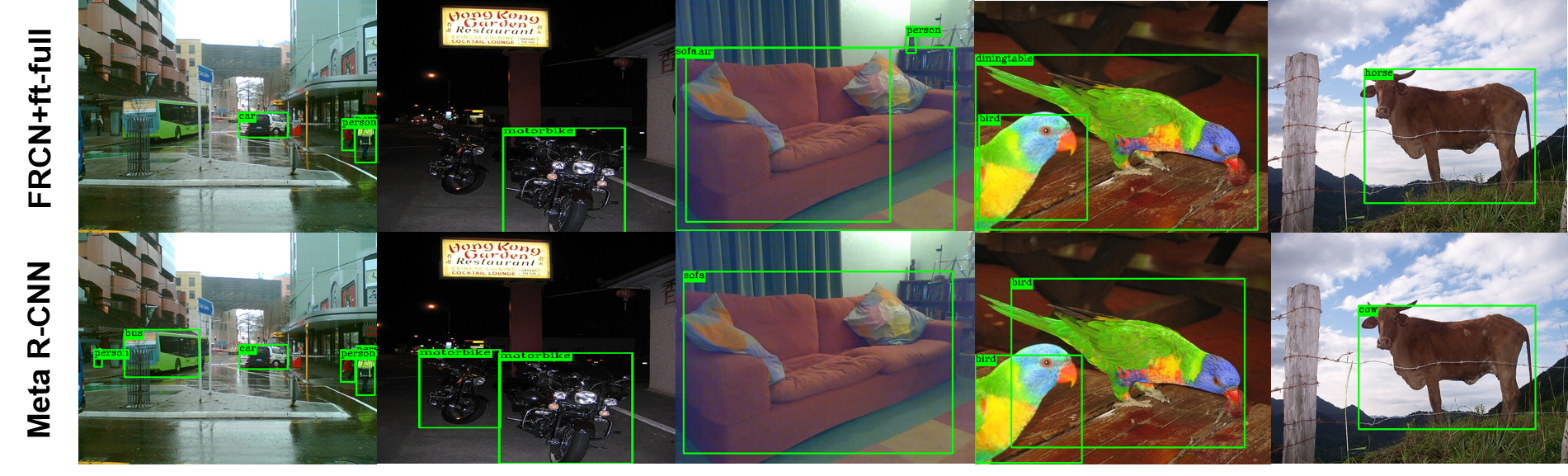}
	\caption{The visualization of novel-class objects detected by FRCN+ft-full and Meta R-CNN. Compared with Meta R-CNN, FRCN+ft-full is inferior: bboxes in the first two columns are missed; in the middle column is duplicate and the classes are wrong in the last two columns.}\label{fig:example}
	\vspace{-5pt}
\end{figure*}
\begin{table*}[!htb]
	\center
	\setlength{\tabcolsep}{2.5pt}
	\fontsize{8}{10}{\selectfont
		\begin{tabular}{c|c|c c c c c c|c c c c c c}
			\hline
			Shot &Baselines &AP &AP$_{50}$ &AP$_{75}$ &AP$_S$ &AP$_M$ &AP$_L$ &AR$_1$ &AR$_{10}$ &AR$_{100}$ &AR$_S$ &AR$_M$ &AR$_L$ \cr\hline
			\multirow{4}*{10}
			&YOLO-Few-shot~\cite{kang2018few} &5.6 &12.3 &4.6 &0.9 &3.5 &10.5 &10.1 &14.3 &14.4 &1.5 &8.4 &{\color{blue}\textbf{28.2}} \\
			&FRCN+ft &1.3 &4.2 &0.4 &0.4 &0.9 &2.1 &5.5 &8.0 &8.0 &2.4 &6.4 &13.0 \\
			&FRCN+ft-full &{\color{blue}\textbf{6.5}} &{\color{blue}\textbf{13.4}} &{\color{blue}\textbf{5.9}} &{\color{blue}\textbf{1.8}} &{\color{blue}\textbf{5.3}} &{\color{blue}\textbf{11.3}} &{\color{red}\textbf{12.6}} &{\color{blue}\textbf{17.7}} &{\color{blue}\textbf{17.8}} &{\color{blue}\textbf{6.5}} &{\color{blue}\textbf{14.4}} &{\color{red}\textbf{28.6}} \\
			&Meta R-CNN (ours) &{\color{red}\textbf{8.7}$\mathbf{^{+2.2}}$} &{\color{red}\textbf{19.1}$\mathbf{^{+5.7}}$} &{\color{red}\textbf{6.6}$\mathbf{^{+0.7}}$} &{\color{red}\textbf{2.3}$\mathbf{^{+0.5}}$} &{\color{red}\textbf{7.7}$\mathbf{^{+2.4}}$} &{\color{red}\textbf{14.0}$\mathbf{^{+2.7}}$} &{\color{red}\textbf{12.6}$\mathbf{^{+0}}$} &{\color{red}\textbf{17.8}$\mathbf{^{+0.1}}$} &{\color{red}\textbf{17.9}$\mathbf{^{+0.1}}$} &{\color{red}\textbf{7.8}$\mathbf{^{+1.3}}$} &{\color{red}\textbf{15.6}$\mathbf{^{+1.2}}$} &27.2$\mathbf{^{-1.4}}$ \\
			\hline
			\multirow{4}*{30}
			&YOLO-Few-shot~\cite{kang2018few} &9.1 &19.0 &7.6 &0.8 &4.9 &16.8 &13.2 &17.7 &17.8 &1.5 &10.4 &{\color{red}\textbf{33.5}} \\
			&FRCN+ft &1.5 &4.8 &0.5 &0.3 &1.8 &2.0 &7.0 &10.1 &10.1 &5.8 &8.3 &13.5 \\
			&FRCN+ft-full &{\color{blue}\textbf{11.1}} &{\color{blue}\textbf{21.6}} &{\color{blue}\textbf{10.3}} &{\color{red}\textbf{2.9}} &{\color{blue}\textbf{8.8}} &{\color{blue}\textbf{18.9}} &{\color{red}\textbf{15.0}} &{\color{blue}\textbf{21.1}} &{\color{blue}\textbf{21.3}} &{\color{red}\textbf{10.1}} &{\color{blue}\textbf{17.9}} &{\color{blue}\textbf{33.2}} \\
			&Meta R-CNN (ours) &{\color{red}\textbf{12.4}$\mathbf{^{+1.3}}$} &{\color{red}\textbf{25.3}$\mathbf{^{+4.3}}$} &{\color{red}\textbf{10.8}$\mathbf{^{+0.5}}$} &{\color{blue}\textbf{2.8}$\mathbf{^{-0.1}}$} &{\color{red}\textbf{11.6}$\mathbf{^{+2.8}}$} &{\color{red}\textbf{19.0}$\mathbf{^{+1.0}}$} &{\color{red}\textbf{15.0}$\mathbf{^{+0}}$} &{\color{red}\textbf{21.4}$\mathbf{^{+0.3}}$} &{\color{red}\textbf{21.7}$\mathbf{^{+0.4}}$} &{\color{blue}\textbf{8.6}$\mathbf{^{-1.5}}$} &{\color{red}\textbf{20.0}$\mathbf{^{+2.1}}$} &32.1$\mathbf{^{-1.4}}$ \\
			\hline
		\end{tabular}}\vspace{-5pt}
		\caption{Few-shot detection performance on COCO minival set for novel classes. We evaluate the performance for different shot examples of novel classes under FRCN pipeline with ResNet-50. {\color{red}RED}/{\color{blue}BLUE} indicate the SOTA/the second best. (Best viewd in color)}\label{tab:coco_detection}
		\vspace{-8pt}
	\end{table*}
	\begin{table}[!htb]
		\vspace{-6pt}\center
		\caption{The ablation of image-level and RoI-level meta-learning}\setlength{\tabcolsep}{2pt}\label{tab:RoIablations}
		\fontsize{8.5}{10}{\selectfont
			\begin{tabular}{c|ccc|ccc}
				\hline
				shot &Method & Base  & Novel\cr\hline
				\multirow{2}*{3} 
				&full-image meta-learning &43.4 &8.1 \\
				&RoI meta-learning &\textbf{64.8} &\textbf{35.0} \\
				\hline
				\multirow{2}*{10}
				{\tiny }			&full-image meta-learning &61.2 &32.0 \\
				&RoI meta-learning  &\textbf{67.9} &\textbf{51.5}\\
				\hline
			\end{tabular}}
			\vspace{-6pt}
		\end{table}

	\begin{table}[htb]
		\center
		\setlength{\tabcolsep}{2pt}
		\fontsize{8.5}{10}{\selectfont
			\begin{tabular}{c|c c c|c c c}
				\hline
				shot & Ablation (1) & Base  & Novel & Ablation (2) & Base  & Novel\cr\hline
				\multirow{2}*{3}
				&meta-learning (w/o) &38.5 &9.0&meta-loss (w/o) &24.2 &57.7\\
				&meta-learning (w) &\textbf{64.8} &\textbf{35.0}&meta-loss (w) &\textbf{35.0} &\textbf{64.8} \\
				\hline
				\multirow{2}*{10}
				&meta-learning (w/o) &56.9 &40.5 &meta-loss (w/o) &46.6 &64.3\\
				&meta-learning (w)  &\textbf{67.9} &\textbf{51.5}&meta-loss (w) &\textbf{51.5} &\textbf{67.9}\\
				\hline
			\end{tabular}}
			\caption{Ablation studies of (1) \textbf{\emph{meta-learning}} and (2) \textbf{\emph{meta-loss}} (mAP on VOC2007 test set for novel classes and base classes of the first base/novel split under FRCN pipeline with ResNet-101) . }\label{tab:ablations}
			\vspace{-16pt}
		\end{table}
		\begin{table*}[!htb]
			\centering
			\setlength{\tabcolsep}{2.5pt}
			{\fontsize{8}{10.5}\selectfont
				\begin{tabular}{c|c|c c c c c c|c c c c c c}
					\hline
					\multicolumn{2}{c|}{}&\multicolumn{6}{c|}{Box}&\multicolumn{6}{c}{Mask}\cr\cline{1-14}
					\hline
					shot &method &AP &AP$_{50}$ &AP$_{75}$ &AP$_S$ &AP$_M$ &AP$_L$ &AP &AP$_{50}$ &AP$_{75}$ &AP$_S$ &AP$_M$ &AP$_L$ \cr\hline
					\multirow{2}*{5}
					&MRCN+ft-full &1.3 &3.0 &1.1 &0.3 &1.1 &2.4 &1.3 &2.7 &1.1 &{\textbf{0.3}} &0.6 &2.2 \\
					&Meta R-CNN (ours) &{\textbf{3.5}$\mathbf{^{+2.2}}$} &{\textbf{9.9}$\mathbf{^{+6.9}}$} &\textbf{1.2}$\mathbf{^{+0.1}}$ &{\textbf{1.2}$\mathbf{^{+0.9}}$} &\textbf{3.9}$\mathbf{^{+2.8}}$ &\textbf{5.8}$\mathbf{^{+3.4}}$ &{\textbf{2.8}$\mathbf{^{+1.5}}$} &{\textbf{6.9}$\mathbf{^{+4.2}}$} &{\textbf{1.7}$\mathbf{^{+0.6}}$} &\textbf{0.3}$\mathbf{^{+0.0}}$ &{\textbf{2.3}$\mathbf{^{+1.7}}$} &{\textbf{4.7}$\mathbf{^{+2.5}}$}  \\
					\hline
					\multirow{2}*{10}
					&MRCN+ft-full &2.5 &5.7 &1.9 &{{2.0}} &{{2.7}} &3.9 &1.9 &4.7 &1.3 &0.2 &{{1.4}} &3.2 \\
					&Meta R-CNN (ours) &{\textbf{5.6}$\mathbf{^{+3.1}}$} &{\textbf{14.2}$\mathbf{^{+8.5}}$} &{{3.0}$\mathbf{^{+1.1}}$} &\textbf{2.0}$\mathbf{^{+0.0}}$ &\textbf{6.6}$\mathbf{^{+3.9}}$ &{\textbf{8.8}$\mathbf{^{+4.9}}$} &{\textbf{4.4}$\mathbf{^{+2.5}}$} &{\textbf{10.6}$\mathbf{^{+5.9}}$} &{\textbf{3.3}$\mathbf{^{+2.0}}$} &{\textbf{0.5}$\mathbf{^{+0.3}}$} &\textbf{3.6}$\mathbf{^{+2.2}}$ &{\textbf{7.2}$\mathbf{^{+4.0}}$}    \\
					\hline
					\multirow{2}*{20}
					&MRCN+ft-full &4.5 &9.8 &{\textbf{3.4}} &{\textbf{2.0}} &4.6 &6.2 &3.7 &8.5 &2.9 &0.3 &2.5 &5.8 \\
					&Meta R-CNN (ours) &{\textbf{6.2}$\mathbf{^{+1.7}}$} &{\textbf{16.6}$\mathbf{^{+6.8}}$} &2.5$\mathbf{^{-0.9}}$ &1.7$\mathbf{^{-0.3}}$ &{\textbf{6.7}$\mathbf{^{+2.1}}$} &{\textbf{9.6}$\mathbf{^{+3.4}}$} &{\textbf{6.4}$\mathbf{^{+2.7}}$} &{\textbf{14.8}$\mathbf{^{+6.3}}$} &{\textbf{4.4}$\mathbf{^{+1.5}}$} &{\textbf{0.7}$\mathbf{^{+0.4}}$} &{\textbf{4.9}$\mathbf{^{+2.4}}$} &{\textbf{9.3}$\mathbf{^{+3.5}}$} \\
					\hline
				\end{tabular}}\caption{Few-shot detection and instance segmentation performance on COCO minival set for novel classes under Mask R-CNN with ResNet-50. The evaluation based on 5/10/20-shot-object in novel classes (More comprehensive results see our supplementary material). }\label{tab:coco_segmentation}
				\vspace{-12pt}
			\end{table*} 

Let's consider detailed evaluation in Table~{\color{red}\ref{tab:sub_split1}} based on the first base/novel-class split. Note that, FRCN+joint achieved SOTA in base classes, however, at the price of the performance disaster in novel classes ($72.7$ in base classes yet $4.3$ in novel classes given $K$=3). This sharp contrast caused by the extreme object quantity imbalance in the few-shot setup, further reveal the fragility of FRCN in the generalization problem. On the other hand, we find that Meta R-CNN outperforms YOLO-Few-shot both in base classes and novel classes, which means that Meta R-CNN is the SOTA few-shot detector. Finally, Meta R-CNN outperforms all other baselines in mAP. This observation is significant: \textbf{Meta R-CNN would not sacrifice the overall performance to make few-shot learning.} In Fig~\ref{fig:example}, we visualize some comparison between FRCN+ft+full and Meta R-CNN on detecting novel-class objects.        

 
		

\textbf{MS COCO.} We evaluate 10-shot /30-shot setups on MS COCO~\cite{lin2014microsoft} benchmark and report the standard COCO metrics. The results on novel classes are presented in  Table~{\color{red}\ref{tab:coco_detection}}. It shows that Meta R-CNN significantly outperforms other baselines and YOLO-Few-shot. Note that, the performance gain is obtained by our method compared to YOLO-Few-shot (12.4\% vs. 11.1\%). The improvement is lower than those on PASCAL VOC, since MS COCO is more challenging with more complex scenarios such as occlusion, ambiguities and small objects.


	\textbf{MS COCO to PASCAL.} In this cross-dataset few-shot object detection setup, all the baselines are trained with 10-shot objects in novel classes on MS COCO while they are evaluated on PASCAL VOC2007 test set. Distinct from the previous experiments that focus on evaluating cross-category model generalization, this setup further to reveal the cross-domain generalization ability. FRCN+ft and FRCN+ft-full get the detection performances of 19.2\% and 31.2\% respectively. The few-shot object detector YOLO-Few-shot obtains 32.3\%. Instead, Meta R-CNN achieves 37.4\%, reaping a significant performance gain (approximately 5\% mAP) against the second best. 
	
	
	\vspace{-2pt}\subsection{Ablation}\label{6.2}\vspace{-4pt}
	Here we conduct comprehensive ablation studies to uncover Meta R-CNN. These ablations are based on 3/10-shot object detection performances on PASCAL VOC in the first base/novel split setup.
	
	
	\textbf{Backbone.} We ablate the backbone (i.e. ResNet-34~\cite{He2017Mask} and ResNet-101~\cite{He2017Mask}) of Meta R-CNN to observe the object detection performances in base and novel classes (Table~{\color{red}\ref{tab:resnet}}). It's observed that our framework significantly outperforms the FRCN-ft-full on base and novel classes across different backbones (large margins of 35.0\% vs. 32.8\% with ResNet-34 and 51.5\% vs. 45.6\% with ResNet-101 on novel classes). These verify the potential of Meta R-CNN that can be flexibly-deployed across different backbones and consistently outperforms the baseline methods.
	
	\textbf{RoI meta-learning.} Since Meta R-CNN is formally devised as a meta-learner, it would be important to observe whether it is truly improved by RoI meta-learning. To verify our claim, we ablate Meta R-CNN from two aspects: 1). using meta-learning or not (Ablation {\color{red}1} in Table~\ref{tab:ablations}); 2). meta-learning on full-image or RoI features (Table~\ref{tab:RoIablations}). As illustrated in  Table~{\color{red}\ref{tab:ablations}} (Ablation {\color{red}1}), meta-learning significantly boosts Meta R-CNN performance by clear large margins both in novel classes (35.0\% vs.9.0\% in 3-shot; 51.5\% vs.40.5\% in 10-shot) and in base classes (38.5\% vs.64.8\% in 3-shot; 67.9\% vs.56.9\% in 10-shot). As $K$ decreases, the improvement will be more significant. In Table~\ref{tab:RoIablations}, we have observed that full-image meta-learning suffers heavy performance drop compared with RoI meta-learning and moreover, it even performs worse than the Faster R-CNN trained without meta-strategy. It shows that RoI meta-learning indeed encourages the generalization of the R-CNN family. 
	
	\textbf{Meta-loss $L_{\rm meta}(\boldsymbol{\phi})$.} Meta R-CNN takes the control of Faster R-CNN by way of class attentive vectors. Their reasonable diversity would lead to the performance improvement when detecting the objects in different classes. To verify our claim, we ablate the meta-loss $L_{\rm meta}(\boldsymbol{\phi})$ used to increase the diversity of class-attentive vectors. The ablation is shown in Table{\color{red}~\ref{tab:ablations}} Ablation {\color{red}2}. Obviously, the Meta R-CNN performances in base and novel classes are significantly improved by adding the meta-loss.

			\vspace{-2pt}\subsection{Few-shot object segmentation}\label{6.3}\vspace{-2pt}
			
			As we demonstrated in our methodology, Meta R-CNN is a versatile meta-learning framework to achieve few-shot object structure prediction, especially, not just limited in the object detection task. To verify our claim, we deploy PRN to change a Mask R-CNN~\cite{He2017Mask} (MRCN) into its Meta R-CNN version. This Meta R-CNN using ResNet-50~\cite{he2016deep} as its backbone, would be evaluated on the instance-level object segmentation track on MS COCO benchmark. We report the standard COCO metrics based on object detection and segmentation. Noted that, AP in object segmentation is evaluated by using \emph{mask} IoU. We use the trainval35k images for training and val5k for testing where the 20 classes in PASCAL VOC~\cite{everingham2010pascal} as novel classes and the remaining 60 categories in COCO~\cite{lin2014microsoft} as base classes. Base classes have abundant labeled samples with instance segmentation while novel classes only have \emph{K}-shot annotated bounding boxes and instance segmentation masks. \emph{K} is set to 5,10 and 20 in our object segmentation experiments.
			
			\textbf{Results.}
			Due to the relatively competitive performances of FRCN+ft+full shown in few-shot object detection, we adopt the same-style training strategy for MRCN, leading to \textbf{MRCN+ft+full} on object detection and instance-level object segmentation results in Table.{\color{red}~\ref{tab:coco_segmentation}}.
			It could be observed that our proposed Meta R-CNN is consistently superior to MRCN+ft+full across 5,10,20-shot settings with significant margins in few-shot object segmentation tasks. For instance, Meta R-CNN achieves a 1.7\% performance improvement (6.2\% vs.4.5\%) on object detection and 2.7\% performance improvement (6.4\% vs.3.7\%) on instance segmentation. These evidences further demonstrate the superiority and universality of our Meta R-CNN presenting. Comprehensive results are found in our supplementary material. 
			
			\vspace{-6pt}\section{Discussion and Future Work}\vspace{-6pt}
Few-shot object detection/ segmentation are very valuable as their successes would lead to an extensive variety of visual tasks generalizing to newly-emerged concepts without heavily consuming labor annotation. Our work takes an insightful step towards the successes by proposing a flexible and simple yet effective framework, \emph{e.g.}, Meta R-CNN. Standing on the shoulders of Faster/ Mask R-CNN, Meta R-CNN overcomes the shared weakness of existing meta-learning algorithms that almost disable to recognize the semantic information entangled with multiple objects. Simultaneously, it endows traditional Faster/ Mask R-CNN with the generalization capability in front of few-shot objects in novel classes. It is lightweight, plug-and-play, and performs impressively in few-shot object detection/ segmentation. It is worth noting that, as Meta R-CNN solely remodels the predictor branches into a meta-learner, it potentially can be extended to a broad range of models \cite{Gkioxari2017Detecting,Hu2017Learning,Kirillov2019Panoptic,Yang2018Graph} in the entire R-CNN family. To this Meta R-CNN might enable visual structure prediction in the more challenging few-shot conditions, \emph{e.g.}, few-shot relationship detection and others.

\section{Appendix}
\subsection{Accelerated task adaptation}
Meta-learning facilitates Faster R-CNN to detect novel-class few-shot objects. Through the lens of stochastic optimization, it gives the credits to the task adaptation acceleration. More specifically, we observe the performance comparison between Faster R-CNN (trained by two-phase strategy, \emph{i.e.}, FRCN+ft-full) and Meta R-CNN over iterations. As shown in Fig \ref{as}, Meta R-CNN presents as an envelope that upper bounds Faster R-CNN. It indicates meta-learning encouraging faster performance improvement to novel-class object detection. 
\begin{figure}[h]\centering
	\includegraphics[width=5cm]{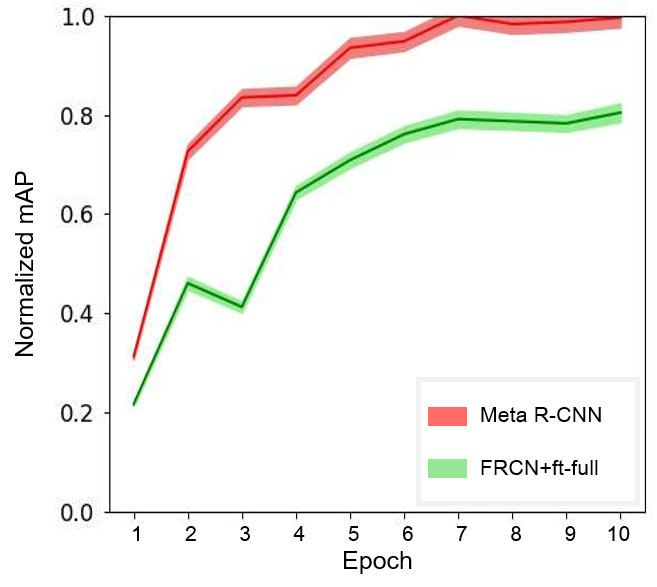}\caption{Normalized mAP \emph{w.r.t.} novel-class object detection over iterations. The mean and variance values of Normalized mAP are computed by class-specific Normalized AP, which is normarlized by the converged value
		of AP against number of
		training iterations.}\vspace{-16pt}
\end{figure}\label{as}

\subsection{Attentive vector analysis}
As we mentioned in the paper, Meta R-CNN takes class attentive vectors to remodel Faster R-CNN, while class attentive vectors are inferred by averaging the object attentive vectors in each class. It implies that learning good representation of object attentive vectors would lead to the success of Meta R-CNN. To this end, we visualize the object attentive vectors used for testing by t-SNE~\cite{maaten2008visualizing}, and compare the same visualization when Meta R-CNN is trained without meta-loss \big($L_{\rm meta}(\boldsymbol{\phi})$\big). All are illustrated in Fig \ref{vis}. First, we find that object attentive vectors tend to cluster together when they belong to the same class and repulse those from the other classes (See Fig \ref{vis} (a)). These object attentive vectors produce more deterministic class attentive vector (less inter-class variance when choosing different objects to induce class attentive vectors). To this Meta R-CNN is endowed with more stable performance, since class attentive vectors would not significant change when objects change. Distinct from this, when Meta R-CNN is trained without meta-loss \big(Fig \ref{vis} (b)\big), object attentive vectors become more diverse and the inter-class variance is very large. These object attentive vectors bring about two negative effects to Meta R-CNN: \textbf{1).} Due to the large inter-class variance, the trained model suffers unstable performances: if we change the objects, the according class attentive vectors will significantly change. \textbf{2).} The inferred class attentive vectors are probably close, resulting ambiguous object detection produced by the corresponding class-specific predictor heads.  

In Fig \ref{vis} (a), it is also observed that the classes with similar semantics would be closer to those with different semantics. For instance, `Car', `Bus', `Train' are close together, as they all belong to vehicle. The observation unveils that Meta R-CNN may achieve novel-class object detection by the aid of the base-class objects that share similar semantic information.  


\begin{figure}[h]\centering
	\vspace{-2pt}\includegraphics[width=8.8cm]{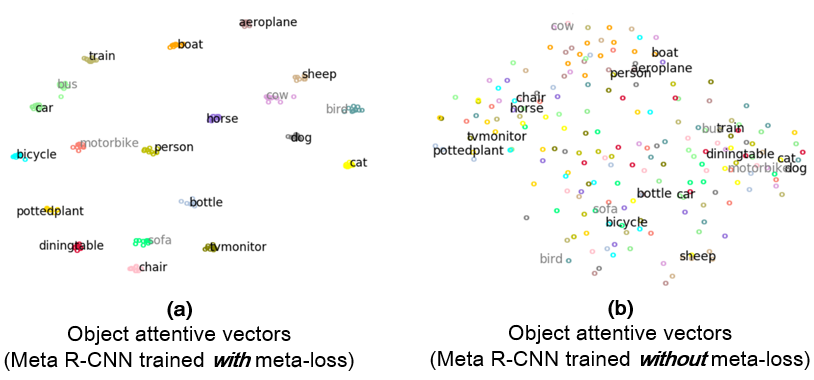}\caption{The t-SNE visualization of object attentive vectors with respect to Meta R-CNN trained w/wo meta-loss. For each class, 10 objects are taken to produce the object attentive vectors for visualization. Color indicates class (Best viewed in color).}\vspace{-4pt}
\end{figure}\label{vis}
\begin{table*}[t]
	\center
	\setlength{\tabcolsep}{1.2pt}
	\fontsize{7.6}{10.5}{\selectfont
		\begin{tabular}{c |c| c c c c c c | c c c c c c |c c c c c c| c c c c|c}
			\hline
			\multicolumn{2}{c|}{}&\multicolumn{6}{c|}{\textbf{Novel-class Split-1} }&\multicolumn{6}{c|}{\textbf{Novel-class Split-2}}&\multicolumn{6}{c|}{\textbf{Novel-class Split-3}}\\
			Shot & Baselines & bird & bus & cow & mbike & sofa & mean &aero &bottle &cow &horse &sofa &mean &boat  &cat  &mbike &sheep &sofa  &mean \\ \hline
			
			\multirow{5}*{1}
			&YOLO-Few-shot\cite{kang2018few} &\color{blue}\textbf{13.5} &10.6 &\color{red}\textbf{31.5} &13.8 &\color{red}\textbf{4.3} &\color{blue}\textbf{14.8} &\color{blue}\textbf{11.8} &\color{red}\textbf{9.1} &15.6 &\color{red}\textbf{23.7} &\color{red}\textbf{18.2} &\color{red}\textbf{15.7}&\color{red}\textbf{10.8} &\color{red}\textbf{44.0} &\color{blue}\textbf{17.8} &\color{red}\textbf{18.1} &\color{red}\textbf{5.3} &\color{red}\textbf{19.2}\\
			&FRCN+joint &9.7 &0.0 &1.5 &0.5 &1.8 &2.7 &1.6 &0.3 &3.2 &3.6 &\color{blue}\textbf{0.8} &1.9 &0.2 &21.9 &0.0 &1.1 &\color{blue}\textbf{3.0} &5.2\\ 
			&FRCN+ft &13.4 &14.8 &4.9 &\color{blue}\textbf{25.6} &0.7 &11.9 &0.5 &0.2 &15.9 &12.2 &0.6 &5.9 &\color{blue}\textbf{10.4} &7.3 &13.1 &3.5 &0.6 &5.0\\
			&FRCN+ft-full &\color{red}\textbf{14.3} &\color{blue}\textbf{16.6} &\color{blue}\textbf{16.4} &18.7 &\color{blue}\textbf{2.9} &13.8 &0.5 &0.4 &\color{blue}\textbf{22.7} &\color{blue}\textbf{15.0} &0.7 &7.9  &0.8 &26.4 &12.3 &9.3 &0.1 &9.8\\
			&Meta R-CNN (ours) &6.1 &\color{red}\textbf{32.8} &15.0 &\color{red}\textbf{35.4} &0.2 &\color{red}\textbf{19.9} &\color{red}\textbf{23.9} &\color{blue}\textbf{0.8} &\color{red}\textbf{23.6} &3.1 &0.7 &\color{blue}\textbf{10.4} &0.6 &\color{blue}\textbf{31.1} &\color{red}\textbf{28.9} &\color{blue}\textbf{11.0} &0.1 &\color{blue}\textbf{14.3}\\
			\hline
			\multirow{5}*{2}
			&YOLO-Few-shot\cite{kang2018few} &\color{red}\textbf{21.2} &12.0 &16.8 &\color{blue}\textbf{17.9} &\color{red}\textbf{9.6} &15.5 &\color{red}\textbf{28.6} &\color{blue}\textbf{0.9} &27.6 &0.0 &\color{red}\textbf{19.5} &\color{blue}\textbf{15.3}&\color{blue}\textbf{5.3} &\color{red}\textbf{46.4} &\color{blue}\textbf{18.4} &\color{red}\textbf{26.1} &\color{red}\textbf{12.4} &\color{red}\textbf{21.7}\\
			&FRCN+joint &12.4 &0.1 &2.2 &0.3 &0.5 &3.1 &2.3 &0.2 &3.9 &5.4 &1.0 &2.6 &1.3 &25.0 &0.2 &9.7 &\color{blue}\textbf{1.5} &7.5\\
			&FRCN+ft &5.4 &19.0 &39.8 &16.6 &1.2 &16.4 &3.6 &\color{red}\textbf{1.3} &13.1 &23.3 &\color{blue}\textbf{1.4} &8.5 &\color{blue}\textbf{5.3} &16.9 &10.2 &14.3 &1.1 &9.6\\
			&FRCN+ft-full &8.1 &\color{blue}\textbf{25.9} &\color{red}\textbf{49.3} &13.0 &\color{blue}\textbf{1.5} &\color{blue}\textbf{19.6} &3.5 &0.1 &\color{blue}\textbf{36.1} &\color{blue}\textbf{35.7} &1.1 &\color{blue}\textbf{15.3} &2.2 &\color{blue}\textbf{25.6} &13.9 &13.9 &0.9 &11.3\\
			&Meta R-CNN (ours) &\color{blue}\textbf{17.2} &\color{red}\textbf{34.4} &\color{blue}\textbf{43.8} &\color{red}\textbf{31.8} &0.4 &\color{red}\textbf{25.5} &\color{blue}\textbf{12.4} &0.1 &\color{red}\textbf{44.4} &\color{red}\textbf{50.1} &0.1 &\color{red}\textbf{19.4} &\color{red}\textbf{10.6} &24.0 &\color{red}\textbf{36.2} &\color{blue}\textbf{19.2} &0.8 &\color{blue}\textbf{18.2}\\
			\hline
			\multirow{5}*{3}
			&YOLO-Few-shot~\cite{kang2018few} &26.1 &19.1 &40.7 &20.4 &\color{red}\textbf{27.1} &26.7 &29.4 &\color{red}\textbf{4.6} &34.9 &6.8 &\color{red}\textbf{37.9} &22.7 &\color{blue}\textbf{11.2} &\color{red}\textbf{39.8} &\color{blue}\textbf{20.9} &23.7 &\color{red}\textbf{33.0} &\color{blue}\textbf{25.7}\\
			&FRCN+joint &13.7 &0.4 &6.4 &0.8 &0.2 &4.3   &16.7 &0.2 &7.4 &15.7 &0.5 &8.1&0.2 &37.2 &0.6 &17.2 &0.1 &11.1\\
			&FRCN+ft &\color{red}\textbf{31.1} &24.9 &\color{blue}\textbf{51.7} &23.5 &13.6 &29.0 &\color{blue}\textbf{29.8} &0.1 &40.3 &43.8 &2.9 &23.4&3.7 &32.8 &18.2 &30.7 &5.0 &18.1\\
			&FRCN+ft-full &29.1 &\color{blue}\textbf{34.1} &\color{red}\textbf{55.9} &\color{blue}\textbf{28.6} &\color{blue}\textbf{16.1} &\color{blue}\textbf{32.8} &\color{red}\textbf{31.9} &\color{blue}\textbf{0.3} &\color{blue}\textbf{45.2} &\color{blue}\textbf{50.4} &3.4 &\color{blue}\textbf{26.2}&10.6 &27.2 &16.5 &\color{blue}\textbf{31.7} &9.5 &19.1\\
			&Meta R-CNN (ours) &\color{blue}\textbf{30.1} &\color{red}\textbf{44.6} &50.8 &\color{red}\textbf{38.8} &10.7 &\color{red}\textbf{35.0} &25.2 &0.1 &\color{red}\textbf{50.7} &\color{red}\textbf{53.2} &\color{blue}\textbf{18.8} &\color{red}\textbf{29.6}&\color{red}\textbf{16.3} &\color{blue}\textbf{39.7} &\color{red}\textbf{32.6} &\color{red}\textbf{38.8} &\color{blue}\textbf{10.3} &\color{red}\textbf{27.5}\\
			\hline
			\multirow{5}*{5}
			&YOLO-Few-shot\cite{kang2018few} &31.5 &21.1 &39.8 &\color{blue}\textbf{40.0} &\color{blue}\textbf{37.0} &33.9 &\color{red}\textbf{33.1} &\color{red}\textbf{9.4} &38.4 &25.4 &\color{red}\textbf{44.0} &30.1 &\color{blue}\textbf{14.2} &\color{red}\textbf{57.3} &\color{blue}\textbf{50.8} &38.9 &\color{blue}\textbf{41.6} &\color{blue}\textbf{40.6}\\
			&FRCN+joint &17.4 &7.9 &9.6 &14.0 &9.1 &11.8  &3.2 &\color{blue}\textbf{4.5} &16.1 &24.8 &0.6 &9.9 &1.6 &39.7 &3.2 &16.4 &3.4 &12.9\\
			&FRCN+ft &31.3 &36.5 &54.1 &26.5 &36.2 &36.9 &17.5 &2.3 &39.6 &\color{blue}\textbf{55.0} &31.2 &29.1 &5.1 &41.7 &33.1 &36.2 &37.9 &30.8\\
			&FRCN+ft-full &\color{red}\textbf{36.1} &\color{blue}\textbf{44.6} &\color{red}\textbf{56.0} &33.5 &\color{red}\textbf{37.2} &\color{blue}\textbf{41.5}  &23.1 &3.9 &\color{blue}\textbf{44.7} &54.0 &32.2 &\color{blue}\textbf{31.6} &11.0 &\color{blue}\textbf{51.8} &36.0 &\color{blue}\textbf{41.3} &34.6 &35.0\\
			&Meta R-CNN (ours) &\color{blue}\textbf{35.8} &\color{red}\textbf{47.9} &\color{blue}\textbf{54.9} &\color{red}\textbf{55.8} &34.0 &\color{red}\textbf{45.7} &\color{blue}\textbf{28.5} &0.3 &\color{red}\textbf{50.4} &\color{red}\textbf{56.7} &\color{blue}\textbf{38.0} &\color{red}\textbf{34.8} &\color{red}\textbf{16.6} &45.8 &\color{red}\textbf{53.9} &\color{red}\textbf{41.5} &\color{red}\textbf{48.1} &\color{red}\textbf{41.2}\\
			\hline
			\multirow{5}*{10}
			&YOLO-Few-shot~\cite{kang2018few} &30.0 &\color{red}\textbf{62.7} &43.2 &\color{red}\textbf{60.6} &39.6 &\color{blue}\textbf{47.2} &43.2 &\color{red}\textbf{13.9} &41.5 &58.1 &39.2 &\color{blue}\textbf{39.2} &\color{red}\textbf{20.1} &51.8 &55.6 &\color{blue}\textbf{42.4} &36.6 &41.3\\
			&FRCN+joint &14.6 &20.3 &19.2 &24.3 &2.2 &16.1 &17.6 &\color{blue}\textbf{9.1} &13.8 &21.6 &0.8 &12.6 &2.3 &43.0 &17.4 &12.6 &1.0 &15.3
			\\
			&FRCN+ft &31.3 &36.5 &\color{red}\textbf{54.1} &26.5 &36.2 &36.9 &\color{blue}\textbf{46.5} &4.5 &34.0 &57.9 &1.1 &28.8 &15.5 &\color{blue}\textbf{65.2} &53.6 &40.9 &41.9 &43.4\\
			&FRCN+ft-full &\color{blue}\textbf{40.1} &47.8 &45.5 &47.5 &\color{red}\textbf{47.0} &45.6 &44.3 &3.0 &\color{blue}\textbf{42.9} &\color{blue}\textbf{59.4} &\color{blue}\textbf{46.2} &39.1&\color{blue}\textbf{19.4} &64.3 &\color{blue}\textbf{57.3} &40.9 &\color{blue}\textbf{43.4} &\color{blue}\textbf{45.1}\\
			&Meta R-CNN (ours) &\color{red}\textbf{52.5} &\color{blue}\textbf{55.9} &\color{blue}\textbf{52.7} &\color{blue}\textbf{54.6} &\color{blue}\textbf{41.6} &\color{red}\textbf{51.5} &\color{red}\textbf{52.8} &3.0 &\color{red}\textbf{52.1} &\color{red}\textbf{70.0} &\color{red}\textbf{49.2} &\color{red}\textbf{45.4} &13.9 &\color{red}\textbf{72.6} &\color{red}\textbf{58.3} &\color{red}\textbf{47.8} &\color{red}\textbf{47.6} &\color{red}\textbf{48.1}\\
			\hline
			\hline
		\end{tabular}}\vspace{5pt}
		\caption{AP and mAP on VOC2007 test set for novel classes and base classes of the first base/novel split. We evaluate the performance for different shots novel-class examples with FRCN under ResNet-101. {\color{red}RED}/{\color{blue}BLUE} indicate the SOTA/the second best. (Best viewd in color)
		}\label{tab:split1}
	\end{table*}
	
	\begin{table*}[!htb]
		\centering
		\setlength{\tabcolsep}{2.5pt}
		{\fontsize{8}{11.5}\selectfont
			\begin{tabular}{c|c|c c c c c c|c c c c c c}
				\hline
				\multicolumn{2}{c|}{COCO \textbf{Novel-class Split-1}}&\multicolumn{6}{c|}{Box}&\multicolumn{6}{c}{Mask}\cr\cline{1-14}
				\hline
				shot &method &AP &AP$_{50}$ &AP$_{75}$ &AP$_S$ &AP$_M$ &AP$_L$ &AP &AP$_{50}$ &AP$_{75}$ &AP$_S$ &AP$_M$ &AP$_L$ \cr\hline
				\multirow{3}*{5}
				&MRCN+ft-full &1.3 &3.0 &1.1 &0.3 &1.1 &2.4 &1.3 &2.7 &1.1 &{\textbf{0.3}} &0.6 &2.2 \\
				&Meta R-CNN (224x224) &{2.4$\mathbf{^{+1.1}}$} &{5.8$\mathbf{^{+2.8}}$} &{\textbf{1.5}$\mathbf{^{+0.4}}$} &{0.8$\mathbf{^{+0.5}}$} &{2.5$\mathbf{^{+1.4}}$} &{3.7$\mathbf{^{+1.3}}$} &{2.2$\mathbf{^{+0.9}}$} &{4.9$\mathbf{^{+2.2}}$} &{\textbf{1.7}$\mathbf{^{+0.6}}$} &0.2$\mathbf{^{-0.1}}$ &{1.7$\mathbf{^{+1.1}}$} &{3.6$\mathbf{^{+1.4}}$}  \\
				&Meta R-CNN (600x600) &{\textbf{3.5}$\mathbf{^{+2.2}}$} &{\textbf{9.9}$\mathbf{^{+6.9}}$} &{1.2}$\mathbf{^{+0.1}}$ &{\textbf{1.2}$\mathbf{^{+0.9}}$} &\textbf{3.9}$\mathbf{^{+2.8}}$ &\textbf{5.8}$\mathbf{^{+3.4}}$ &{\textbf{2.8}$\mathbf{^{+1.5}}$} &{\textbf{6.9}$\mathbf{^{+4.2}}$} &{\textbf{1.7}$\mathbf{^{+0.6}}$} &\textbf{0.3}$\mathbf{^{+0.0}}$ &{\textbf{2.3}$\mathbf{^{+1.7}}$} &{\textbf{4.7}$\mathbf{^{+2.5}}$}  \\
				\hline
				\multirow{3}*{10}
				&MRCN+ft-full &2.5 &5.7 &1.9 &{\textbf{2.0}} &{{2.7}} &3.9 &1.9 &4.7 &1.3 &0.2 &{1.4} &3.2 \\
				&Meta R-CNN (224x224) &{4.3$\mathbf{^{+1.8}}$} &{9.4$\mathbf{^{+3.7}}$} &{\textbf{3.3}$\mathbf{^{+1.4}}$} &1.3$\mathbf{^{-0.7}}$ &0.4$\mathbf{^{-2.3}}$ &{{6.4}$\mathbf{^{+2.5}}$} &{{3.7}$\mathbf{^{+1.8}}$} &{{8.4}$\mathbf{^{+3.7}}$} &{{2.9}$\mathbf{^{+1.6}}$} &{{0.3}$\mathbf{^{+0.1}}$} &0.2$\mathbf{^{-1.2}}$ &{{5.6}$\mathbf{^{+2.4}}$}  \\
				&Meta R-CNN (600x600) &{\textbf{5.6}$\mathbf{^{+3.1}}$} &{\textbf{14.2}$\mathbf{^{+8.5}}$} &{{3.0}$\mathbf{^{+1.1}}$} &\textbf{2.0}$\mathbf{^{+0.0}}$ &\textbf{6.6}$\mathbf{^{+3.9}}$ &{\textbf{8.8}$\mathbf{^{+4.9}}$} &{\textbf{4.4}$\mathbf{^{+2.5}}$} &{\textbf{10.6}$\mathbf{^{+5.9}}$} &{\textbf{3.3}$\mathbf{^{+2.0}}$} &{\textbf{0.5}$\mathbf{^{+0.3}}$} &\textbf{3.6}$\mathbf{^{+2.2}}$ &{\textbf{7.2}$\mathbf{^{+4.0}}$}  \\
				\hline
				\multirow{2}*{20}
				&MRCN+ft-full &4.5 &9.8 &{\textbf{3.4}} &{\textbf{2.0}} &4.6 &6.2 &3.7 &8.5 &2.9 &0.3 &2.5 &5.8 \\
				&Meta R-CNN (224x224) &{\textbf{6.2}$\mathbf{^{+1.7}}$} &{\textbf{16.6}$\mathbf{^{+6.8}}$} &2.5$\mathbf{^{-0.9}}$ &1.7$\mathbf{^{-0.3}}$ &{\textbf{6.7}$\mathbf{^{+2.1}}$} &{\textbf{9.6}$\mathbf{^{+3.4}}$} &{\textbf{6.4}$\mathbf{^{+2.7}}$} &{\textbf{14.8}$\mathbf{^{+6.3}}$} &{\textbf{4.4}$\mathbf{^{+1.5}}$} &{\textbf{0.7}$\mathbf{^{+0.4}}$} &{\textbf{4.9}$\mathbf{^{+2.4}}$} &{\textbf{9.3}$\mathbf{^{+3.5}}$} \\
				\hline
			\end{tabular}}\vspace{5pt}\caption{Few-shot detection and instance segmentation performance on COCO minival set for novel classes under Mask R-CNN with ResNet-50. The evaluation based on 5/10/20-shot-object in novel classes. }\label{tab:coco_segmentation_1}
			\end{table*} 	 		
			
			\begin{table*}[!htb]
				\centering
					\setlength{\tabcolsep}{2.5pt}
					{\fontsize{8}{11.5}\selectfont
						\begin{tabular}{c|c|c c c c c c|c c c c c c}
							\hline
							\multicolumn{2}{c|}{COCO \textbf{Novel-class Split-2}}&\multicolumn{6}{c|}{Box}&\multicolumn{6}{c}{Mask}\cr\cline{1-14}
							\hline
							shot &method &AP &AP$_{50}$ &AP$_{75}$ &AP$_S$ &AP$_M$ &AP$_L$ &AP &AP$_{50}$ &AP$_{75}$ &AP$_S$ &AP$_M$ &AP$_L$ \cr\hline
							\multirow{2}*{5}
							&MRCN+ft-full &2.3 &4.4  &\textbf{2.3}  &0.6 & 2.3 & 3.2 & 2.1  & 3.9  & \textbf{2.0} &  0.3 &  1.8 &  3.1 \\
							&Meta R-CNN (224x224) &\textbf{3.3}$\mathbf{^{+1.0}}$ & \textbf{9.4}$\mathbf{^{+5.0}}$  & 1.1$\mathbf{^{-1.2}}$  & \textbf{1.7}$\mathbf{^{+1.1}}$ &  \textbf{3.9}$\mathbf{^{+1.6}}$  &4.4$\mathbf{^{+1.2}}$&  \textbf{ 2.3}$\mathbf{^{+0.2}}$   &\textbf{5.1 }$\mathbf{^{+1.2}}$ & 1.8$\mathbf{^{-0.2}}$  & \textbf{0.4}$\mathbf{^{+0.1}}$   &\textbf{2.2}$\mathbf{^{+0.4}}$  & \textbf{3.8}$\mathbf{^{+0.7}}$  \\
							&Meta R-CNN (600x600) &3.1$\mathbf{^{+0.8}}$ & 8.9$\mathbf{^{+4.5}}$  & 1.1$\mathbf{^{-1.2}}$  & 1.1$\mathbf{^{+0.6}}$ &  3.0$\mathbf{^{+0.7}}$  &\textbf{5.1}$\mathbf{^{+1.9}}$&  2.2$\mathbf{^{+0.1}}$   &4.7$\mathbf{^{+0.8}}$ & 1.9$\mathbf{^{-0.1}}$  & \textbf{0.4}$\mathbf{^{+0.1}}$   &1.7$\mathbf{^{-0.1}}$  & 3.2$\mathbf{^{+0.1}}$  \\
							\hline
							\multirow{2}*{10}
							&MRCN+ft-full &2.6   &6.0   &\textbf{1.8}   &1.2 &  2.7 &  3.6 &  2.8  & 5.7  &\textbf{ 2.3}  &\textbf{ 0.5}  & 2.6  & 4.1 \\
							&Meta R-CNN (224x224) &\textbf{3.9}$\mathbf{^{+1.3}}$   &\textbf{11.2}$\mathbf{^{+5.2}}$  &1.4$\mathbf{^{-0.4}}$   &\textbf{1.9}$\mathbf{^{+0.7}}$  & \textbf{4.0}$\mathbf{^{+1.3}}$ & 5.9 $\mathbf{^{+2.3}}$&  \textbf{2.9}$\mathbf{^{+0.1}}$   &6.3$\mathbf{^{+0.6}}$ & 2.1$\mathbf{^{-0.2}}$  & \textbf{0.5}$\mathbf{^{+0.0}}$   &\textbf{2.8}$\mathbf{^{+0.2}}$ &  \textbf{5.0}$\mathbf{^{+0.9}}$ \\
							&Meta R-CNN (600x600) &\textbf{3.9}$\mathbf{^{+1.3}}$ & 11.0$\mathbf{^{+5.0}}$  & 1.7$\mathbf{^{-0.1}}$  & 1.7$\mathbf{^{+0.5}}$ &  3.9$\mathbf{^{+1.2}}$  &\textbf{6.2}$\mathbf{^{+2.6}}$&  2.8$\mathbf{^{+0.0}}$   &\textbf{6.4}$\mathbf{^{+0.7}}$ & 2.1$\mathbf{^{-0.2}}$  & \textbf{0.5}$\mathbf{^{+0.0}}$   &2.7$\mathbf{^{+0.1}}$  & 4.5$\mathbf{^{+0.4}}$  \\
							\hline
							\multirow{2}*{20}
							&MRCN+ft-full &3.4 &  8.1  & 2.3 &  2.2  & 3.7 &  4.9  & 3.3 &  7.4   &2.3 &  0.8  & 3.2   &5.5 \\
							&Meta R-CNN (ours) &\textbf{4.7}$\mathbf{^{+1.3}}$   & \textbf{10.2}$\mathbf{^{+2.1}}$ & \textbf{3.8}$\mathbf{^{+1.5}}$ &  \textbf{2.8 }$\mathbf{^{+0.6}}$  &\textbf{5.4}$\mathbf{^{+1.7}}$  &\textbf{ 7.2}$\mathbf{^{+2.3}}$   &\textbf{4.5}$\mathbf{^{+1.2}}$   &\textbf{9.4}$\mathbf{^{+2.0}}$   &\textbf{3.8}$\mathbf{^{+1.5}}$   &\textbf{1.1}$\mathbf{^{+0.3}}$  & \textbf{4.5}$\mathbf{^{+1,3}}$   &\textbf{7.8}$\mathbf{^{+2.3}}$ \\
							\hline
						\end{tabular}}\vspace{5pt}\caption{Few-shot detection and instance segmentation performance on COCO minival set for novel classes under Mask R-CNN with ResNet-50. The evaluation based on 5/10/20-shot-object in novel classes. }\label{tab:coco_segmentation_2}
						\vspace{-8pt}
					\end{table*} 	
			
			\vspace{-0pt}
			\subsection{Few-shot object detection}
			\vspace{-0pt}
			In Table \ref{tab:split1}, we conduct the PASCAL VOC experimental results based on few-shot object detection in details. These experiments are based on three different novel / base-class split settings: \textbf{Novel-class Split-1} \emph{(``bird'',
				``bus'',
				``cow'', ``mbike'', ``sofa''/ rest)}; \textbf{Novel-class Split-2} \emph{(``aero'', ``bottle'',``cow'',``horse'',``sofa'' / rest)} and \textbf{Novel-class Split-3} \emph{(``boat'',
				``cat'',
				``mbike'',``sheep'', ``sofa''/ rest)}.    
			
			\subsection{Few-shot object segmentation}
			
			In Table \ref{tab:coco_segmentation_1} \ref{tab:coco_segmentation_2}, we conduct the COCO experiments based on few-shot object segmentation in two different novel/base-class split settings. In novel-class split-1, the novel class selection follows the classes in PASCAL VOC. In novel-class split-2, we randomly choose ('person','car', 'motorcycle', 'airplane', 'bus', 'train', 'cow','elephant','zebra','tennis racket','bed', 'refrigerator','pizza', 'toilet','microwave','truck','umbrella', 'handbag', 'parking meter', 'teddy bear') as novel classes. In the 5-/10-shot experiment in Split-1, we develop two variants from our Meta R-CNN, \emph{i.e.}, (224x224) and (600x600). They indicate different resolution of the input in meta (reference)-set $D_{\rm meta}$. Since object segmentation concerns more detailed semantic than object detection, increasing the resolution of reference image can significantly improve the segmentation performance on those objects in the data-starve categories. For a fair comparison with other baselines, the images used for training ($\mathcal{D}_{\rm train}$) and evaluation ($\mathcal{D}_{\rm test}$) are consistent in 224x224 across all baselines.  
			
			\subsection{Construction Ablation of PRN}
			We additionally test four designs to model a predictor head in different manners: \textbf{concate} (Concatenate the class attentive vector and RoI feature for the class-specific prediction), \textbf{plus} (elementwise-plus of class attentive feature and RoI feature for the class-specific prediction), \textbf{unshare} (The parameters of PRN and R-CNN counterpart are not shared), \textbf{limited meta set} (Only use the image-related classes to generate $D_{\rm meta}$). Results are shown in Table.\ref{tab:ablation1}. \begin{table}[htb]
				\center\caption{The ablation of different variations on PRN}\vspace{-4pt}
				\setlength{\tabcolsep}{2pt}
				\fontsize{8.5}{9.5}{\selectfont
					\begin{tabular}{c|c c c||c| c c c}
						\hline
						shot & Variations & Base  & Novel &shot & Variations & Base  & Novel\cr\hline
						\multirow{5}*{3}
						& concate &\textbf{67.0} &33.6 &\multirow{5}*{10} & concate &\textbf{68.4} &50.5\\
						& plus &64.1  &32.9 && plus &67.9 &48.7 \\
						& unshare &59.8 &21.2 && unshare &67.3 &40.5\\
						& limited meta set &55.8 &33.4 && limited meta set &61.4 &49.9 \\
						& ours &64.8 &\textbf{35.0} &&ours &67.9 &\textbf{51.5}\\
						\hline
					\end{tabular}}
					\label{tab:ablation1}
					\vspace{-6pt}
				\end{table}\textbf{concate} shows superior in ``Base'' object detection while \textbf{ours} (channel-wise attention) performs better in ''Novel'' object detection.

{\small
\bibliographystyle{ieee}
\bibliography{egbib}
}

\end{document}